\documentclass[9.5pt,journal,final,finalsubmission,twocolumn]{IEEEtran}
\usepackage{times}
\usepackage{epsfig}
\usepackage{graphicx}
\usepackage{amsmath,amssymb} 
\usepackage{mathtools}
\usepackage{multirow}
\usepackage{subfig}
\usepackage{color}
\usepackage{rotating}
\usepackage{array}
\usepackage{cite}
\usepackage{url}
\usepackage{booktabs}
\usepackage[ruled]{algorithm}
\usepackage {algorithmic}

 \newcounter{proposition}
 \newcommand{\proposition}{\refstepcounter{proposition} \vskip0.2cm{\bf Proposition I:~~}}

\graphicspath{{./figures/}}

\begin{document}
%
\title{Fusion of Range and Stereo Data for \\High-Resolution Scene-Modeling\thanks{This work has received funding from Agence Nationale de la Recherche under the MIXCAM project number ANR-13-BS02-0010-01.}}
%
%

\author{Georgios~D.~Evangelidis,~Miles Hansard, and~Radu Horaud
\thanks{}
\thanks{G. D. Evangelidis and R. Horaud are with Perception Team, INRIA Grenoble Rh\^one-Alpes, 655, avenue de l'Europe, 38330 Montbonnot Saint-Martin, France, email:georgios.evangelidis@inria.fr, radu.horaud@inria.fr}
\thanks{M. Hansard is with the Vision Group, School of Electronic Engineering and Computer Science, Queen Mary, University of London, Mile End Road,
  London E1 4NS, UK, e-mail:miles.hansard@qmul.ac.uk}

}

\maketitle

\begin{abstract}

This paper addresses the problem of range-stereo fusion, for the construction of high-resolution depth maps. In particular, we combine low-resolution depth data with high-resolution stereo data, in a maximum a posteriori (MAP) formulation. Unlike existing schemes that build on MRF optimizers, we infer the disparity map from a series of local energy minimization problems that are solved hierarchically, by growing sparse initial disparities obtained from the depth data. The accuracy of the method is not compromised, owing to three properties of the data-term in the energy function. Firstly, it incorporates a new correlation function that is capable of providing refined correlations and disparities, via subpixel correction. Secondly, the correlation scores rely on an adaptive cost aggregation step, based on the depth data. Thirdly, the stereo and depth likelihoods are adaptively fused, based on the scene texture and camera geometry. These properties lead to a more selective growing process which, unlike previous seed-growing methods, avoids the tendency to propagate incorrect disparities. The proposed method gives rise to an intrinsically efficient algorithm, which runs at 3FPS on 2.0MP images on a standard desktop computer. The strong performance of the new method is established both by quantitative comparisons with state-of-the-art methods, and by qualitative comparisons using real depth-stereo data-sets.

\end{abstract}


\section{Introduction}\label{sec:intro}

\begin{figure*}[t]
\includegraphics[width=\textwidth]{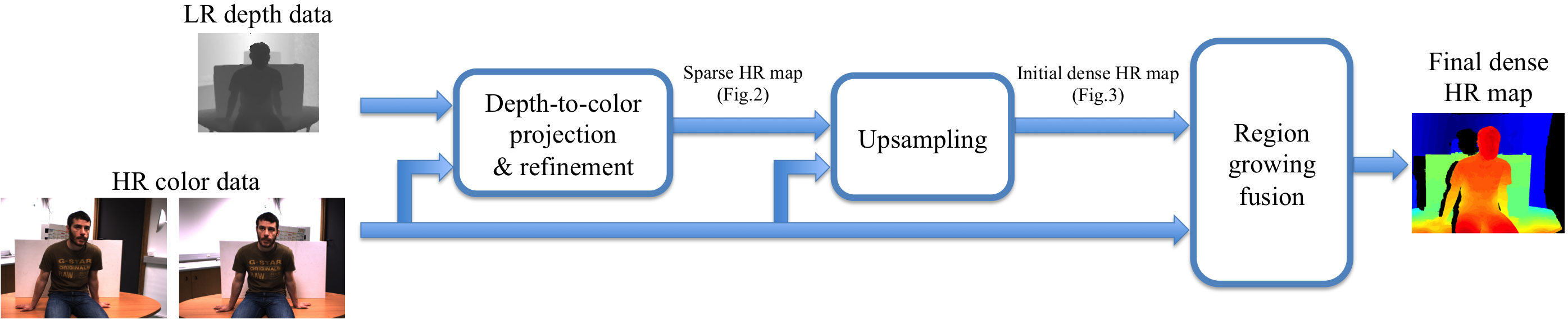}
\caption{
The pipeline of the proposed depth-stereo fusion method. The low-resolution (LR) depth data are projected  onto the color data and refined to yield a high-resolution (HR) sparse disparity map. Starting from these disparity \textit{seeds}, an upsampling process provides an initial HR dense disparity map. Both the HR seeds and the initial dense disparity map are then used by the region-growing depth-stereo fusion to produce the final HR depth map. A prominent feature of our method is that fusion takes place at several data processing stages.
}
\label{fig:proposed_pipeline}
\end{figure*}


Many computer vision methodologies, including dense 3D reconstruction \cite{Henry12,Kim09},
gesture recognition \cite{Bergh11,Evangelidis-ECCVW-2014}, and object detection \cite{Stuckler10} have benefited from recently-developed depth sensors.
These sensors rely on active-light principles, including modulated-light and pulsed-light cameras, commonly denoted time-of-flight (TOF) \cite{HansardBook2012,remondino2013tof}, or projected-pattern triangulation cameras \cite{geng2011structured}. Regardless of the working principle, however, these sensors provide low-resolution (LR) or mid-resolution depth maps that are inadequate for a number of applications such as 3DTV and film production. For example, many tasks in the film production industry can greatly benefit from a high-resolution (HR) and high-quality depth map~\cite{Nair2013}.

While HR depth maps can be obtained from multiple-view matching and reconstruction using standard color cameras, it is well known that stereo matching is problematic when the scene contains weakly textured areas, repetitive patterns, or occlusions; these situations are very common in both indoor and outdoor environments. Active-light sensors do not suffer from these limitations, although their own depth data are quite noisy in the presence of scattering, non-Lambertian materials, and slanted surfaces. The complementary nature of HR stereo and LR depth sensors leads to the design of \textit{mixed} camera systems~\cite{Gudmundsson2008,Hahne2008,Zhu08,Zhu11,Fischer11,Gandhi12,Ruhl2012,Nair2012,Mutto2012locallyConsistent}, which seem to be the most promising approach, at present, for high-quality 3D depth maps. 

In this context, this paper addresses the problem of HR 3D reconstruction from the combination of a photometric camera pair and an active-light camera, provided that the multiple-camera setup is calibrated~\cite{Hansard2014}. The combination of a stereo matching algorithm and of an active-light sensor raises the central question of devising a matching algorithm with the following features: (i)~it considerably increases the resolution of the depth data, e.g., by a factor of ten, (ii)~it eliminates depth-sensor errors wherever possible, (iii)~it overcomes the limitations of stereo algorithms in textureless areas, and (iv)~it is able to compete with a depth sensor in terms of speed. Hence, the availability of an efficient and robust stereo algorithm that takes advantage of LR depth sensors and that provides dense and accurate HR depth maps, possibly with subpixel accuracy, is particularly desirable.

To this end, we propose a 3D reconstruction method that merges depth-sensor measurements with photo-consistency stereo matching. We address the problem from the perspective of seed-growing, starting from a small number of \textit{control points} whose disparities are then propagated to yield a dense disparity map. We show that this can be cast into maximum a posteriori (MAP) formulation (Sec.~\ref{sec:problem_formulation}), which leads, in turn, to a series of local optimization
 problems that are solved hierarchically by a novel region-growing process (Sec.~\ref{sec:fusion}). While the proposed method may not reach the global optimum, it allows us to devise an intrinsically efficient methodology that bridges the gap between global optimizers based on Markov random fields (MRF) and locally-optimal winner-take-all (WTA) strategies (Sec~\ref{sec:discussion_comparison}).

Efficient stereo-only or stereo-depth fusion methods rely on control points, by exploiting either feature correspondences in stereo~\cite{Cech07,Geiger10} or depth data in fusion~\cite{Fischer11,Gandhi12}, and they assume that these points are of very good quality. 
A key contribution of this paper is that this requirement is relaxed in order to devise a method tolerant to bad control points. We propose to truly combine LR depth-sensor data with HR rich photometric information, whenever and wherever possible, showing that fusion is helpful, even in the early stage of depth initialization (Sec.~\ref{sec:sparse_depth_refinement}~\&~\ref{sec:depth_upsampling}). The data term of the proposed MAP formulation benefits from a new cross-correlation function, which provides real-valued disparities via subpixel corrections computed in closed-form (Sec.\ref{sec:subpixel_disparity}), and takes advantage of a depth-guided cross-product aggregation (Sec.~\ref{sec:adaptive_aggregation}). Moreover, the data-term adaptively merges the stereo- and depth-consistency terms guided by the scene texture and the camera geometry (Sec.~\ref{sec:adaptive_fusion}). These advantages lead to a more selective growing-of-disparities process, thus preventing the algorithm from propagating erroneous depth-sensor data -- a phenomenon that is often associated with propagation techniques -- as experimentally verified (Sec.~\ref{sec:experiments}). It is important to note that the proposed method can be used `as is' with any depth sensor, as it requires neither sensor-dependent confidence maps nor a sensor-dependent model. The proposed fusion pipeline is illustrated in Fig.~\ref{fig:proposed_pipeline}.

Supplementary materials, in particular image datasets and Matlab code are available online.\footnote{\url{https://team.inria.fr/perception/research/dsfusion/}}


\section{Related Work}\label{sec:related_work}

We review pure stereo matching methods
with emphasis on local algorithms, owing to their computational suitability for HR images. We also review upsampling methods and depth-stereo fusion methods. More detailed surveys of stereo matching algorithms and depth-stereo fusion methods can be found in \cite{Scharstein02} and \cite{Nair2013}, respectively.



\subsection{Pure Stereo Matching}

Stereo matching methods can be broadly classified into global and local~\cite{Scharstein02}. Global algorithms~\cite{Szeliski2008} typically adopt an MRF formulation and solve a single optimization problem based on a MAP criterion. Despite their superiority over local methods, global algorithms are extremely time-consuming, and hence unattractive for fast fusing of depth-sensor data with high-resolution images. Local algorithms solve per-pixel optimization problems and the state-of-the-art methods build on adaptive cost aggregation~\cite{Yoon06,Rhemann2011fastCost,DeMaezu2011linear,Yang2012_nonLocaLCostAggregation}. Most methods, however, must visit the entire cost volume to find an optimal disparity value at each pixel. This volume grows rapidly with respect to the input, as the width of the image typically multiplies the number of disparities. 
Therefore, although they are able to provide LR disparity maps in real-time, they remain slow and subject to memory issues in HR stereo. Note that global algorithms need several approximations to obtain LR disparity maps in real-time~\cite{Yang2010ConstantSpaceBP}. 

More interestingly, algorithms that rely on control-point correspondences~\cite{Bobick99} are drastically more efficient since they avoid visiting the whole cost volume. Region-growing approaches start from reliable but sparse correspondences (seeds) and propagate them in textured areas~\cite{Cech07}. \cite{Bleyer2011} suggests a similar propagation scheme where orientation-consistent disparities are propagated to neighbors at the cost of finding a plane equation per pixel. \cite{Geiger10} proposed a generative model, where the prior disparity comes from a 2D triangulated mesh whose vertices (control points) are obtained from matches between low-level features. As with~\cite{Cech07}, textureless areas remain intractable and the final map is reliable only when the matches are dense and uniformly distributed over the images. Notice that the proposed method relies on the idea of control points that are transferred from a depth sensor, thus avoiding the limitation owing to untextured areas.

To obtain continuous disparities that are required in many scenarios, e.g., 3D reconstruction~\cite{Yang2007spatial}, local stereo algorithms typically employ two strategies: (i)~fitting a curve around the correlation peak~\cite{Scharstein02,Fischer11} or (ii)~integrating an intensity interpolation kernel into the (dis)similarity function~\cite{Psarakis05} whose optimization leads to subpixel correction. The latter is also the case in the fusion framework of~\cite{Nair2012,Ruhl2012} that inherently takes advantage of inter-pixel depth estimations. 

\subsection{Depth Upsampling and Depth-Stereo Fusion}

Any prior depth information, even at low resolution, is likely to help dense disparity estimation. Apart from a naive interpolation, the bilateral filter~\cite{Tomasi98} can post-process an interpolated map using color HR images for guidance~\cite{Yang2007spatial}. Alternatively, a joint bilateral filter applies spatial and range kernels to the LR depth map and HR color image respectively, so that upsampling is a by-product of filtering~\cite{Kopf2007jointBilateral}. Upsampling methods, however, are limited to clean depth LR data and cannot reconstruct accurate HR maps when LR data are delivered by depth sensors.

The above limitation of the upsampling methods gives rise to fusion approaches that can merge depth and stereo data in either early or late stages, once the sensors are calibrated. Late fusion suggests merging two depth maps, one obtained with stereo and one from the range sensor, possibly upsampled~\cite{Kuhnert2006,Mutto2012locallyConsistent}. The majority of fusion methods, however, merge the stereo and range-sensor data at an earlier processing level.~\cite{Gudmundsson2008} estimates a LR TOF-based disparity map, which initializes a coarse-to-fine stereo scheme~\cite{vanbergen:2002}. A semi-global dynamic programming approach is followed in~\cite{Fischer11} with the TOF-based disparities, wherever available, being considered as error-free matches. In a global framework,~\cite{Hahne2008} produces mid-resolution depth maps by merging TOF and stereo data within a graph-cut stereo-matching paradigm; each energy term exploits both modalities. Likewise, MRF-based formulations have been proposed~\cite{Wang11,Zhu08,Zhu11}. In~\cite{Wang11}, ground control points reflect an extra regularization term in the MRF energy function. The work described in~\cite{Zhu08} uses an MRF scheme to merge depth distributions of each sensor alone, but the goal is a LR depth map.~\cite{Zhu11} extends~\cite{Zhu08} by means of weighted fusion. A balanced fusion (based on several confidence maps) within a total variation model is also proposed in~\cite{Nair2012}. A similar variational model that infers the HR map in a coarse-to-fine manner is adopted by~\cite{Ruhl2012}. Note that, unlike most fusion methods and similar to the proposed one,~\cite{Ruhl2012} is not tuned to a specific depth sensor.

More closely to the present work,~\cite{Gandhi12} fuses the data within the seed-growing method of~\cite{Cech07}. In particular, TOF-based disparities constitute seeds while a triangulation-based interpolated TOF map regularizes the seed-growing process. When the TOF data are noisy, however, this approach tends to produce incorrect disparities, and to propagate false positives during the growing process. The proposed method differs considerably from~\cite{Gandhi12} in terms of depth initialization, cost function, and fusion strategy. The proposed initial map is robust to depth discontinuities while it also guides the cost aggregation inside a window. Moreover, our likelihood term integrates functions that are capable of providing sub-pixel corrections~\cite{Psarakis05}. This turns out to be very beneficial, not only for the continuous nature of the final map, but also for the growing process itself, thus propagating high-confidence disparities only. Note that the subpixel disparity correction is obtained from a closed-form solution -- an interesting feature for efficiency. Finally, our algorithm benefits from an adaptive fusion scheme that better balances the contribution of each modality (depth or color), and that results in fewer unmatched pixels.


\section{Problem Formulation}\label{sec:problem_formulation}

The main mathematical notations that are used throughout the paper 
are summarized in~Table~\ref{table:notations}.
\begin{table}
\centering
\caption{The main mathematical notations used in the paper.}
\begin{tabular}{rp{6.0cm}}
\hline
$p$, $q$: & Pixel locations of the high-resolution grid\\
\mbox{$p$$\downarrow$}, \mbox{$q$$\downarrow$}: & Pixel locations of the sparse grid\\
$d_p$: & Unknown disparity of pixel $p$, initialized by $d_p^0$\\
$d_{p\downarrow}$: & Known disparity of pixel \mbox{$p$$\downarrow$} (observed)\\
$D$: & Unknown HR disparity map, initialized by $D^0$\\
\mbox{$D$$\downarrow$}: & Known sparse version of $D$ (observed)\\
$\mathcal{D}, \mathcal{D}^0$, \mbox{$\mathcal{D}$$\downarrow$}: & Sets of all random variables (disparities) associated with $D$, $D^0$ and \mbox{$D$$\downarrow$} respectively, with $d_p\in\mathcal{D}$, $d_p^0\in\mathcal{D}^0$ and $d_{p\downarrow}\in$\mbox{$\mathcal{D}$$\downarrow$}\\
$t_p$: & Subpixel disparity correction of $p$\\
$\mu_p=(d_p;t_p)$: & Disparity-correction pair referred here to as \emph{meta-disparity} with $|t_p|<1$ \\
$\mathcal{M}=\{\mu_p\}$: & Set of meta-disparities\\
$\mathcal{S}=\{s_p\}$: & Set of observed stereo pixel intensities\\ 
$I_p$: & Intensity of pixel $p$\\
$\mathbf{u}(x,y)$: & a vectorized (zero-mean) form of an intensity window centered at the 2D position $(x,y)$ \\
$\mathcal{N}_p$: & Neighborhood of pixel $p$\\
$\text{med}$: & $2$-d median operator\\
 $I_R$, $D_R$: &  Intensity and disparity maps defined on a sub-region $R$\\
 $E_S(d_p)$, $E_D(d_p)$: & Stereo-based and depth-based energy of $p$ for given disparity\\
 $g(\xi;\gamma)\!=\!e^{-|\xi|/\gamma}$: & Exponential mapping of $\xi$ with scale $\gamma$\\
 \hline
\end{tabular}
\label{table:notations}
\end{table}
As discussed, the direct upsampling of LR depth data suffers from limitations, in particular when the upsampling factor is high.\footnote{In our experiments, the upsampling factor is $10\times$ in each dimension, that is $100\times$ in the number of pixels, e.g., from 0.02Mp  to 2MP.} Therefore, our goal is to build $D$ by jointly taking advantage of both sensing modalities. Given a proper calibration, (e.g.,~\cite{Hansard11,Hansard2014}), the mapping of the LR depth image onto the rectified HR color images will typically yield a \emph{sparse} disparity map \mbox{$D$$\downarrow$}, which is almost evenly distributed across the HR grid. Note that in \cite{Geiger10,Cech07}, the initial matches between the stereo images do not correspond to a uniform sparse version of $D$, as they are unpredictably distributed, due to the reliability of properly detecting interest points in images. Instead, the sparse map obtained with a depth sensor can be used to guide a stereo algorithm, \textit{regardless of the presence or absence of scene texture}.

We propose to model the estimation of $\mathcal{D}$, and therefore the map $D$, as a \emph{maximum a posteriori} (MAP) problem, based on the available depth and stereo observations. However, instead of immediately using \mbox{$\mathcal{D}$$\downarrow$}, we first estimate a dense initial map $D^0$ and its associated set $\mathcal{D}^0$. Then, we obtain the final disparity map by solving the following optimization problem:
\begin{equation}\label{posterior_distribution}
\mathcal{D}^* = \arg\max_{\mathcal{D}} P(\mathcal{D}|\mathcal{S},\mathcal{D}^0),
\end{equation}
where $P(\mathcal{D}|\mathcal{S},\mathcal{D}^0)$ is the posterior distribution of disparities given the observations $\mathcal{S}$ and $\mathcal{D}^0$. 
The proposed depth initialization method is described in Sec.~\ref{sec:depth_initialization} and the proposed solution to the MAP formulation (\ref{posterior_distribution}) is described in Sec.~\ref{sec:fusion}. Note that a reliable estimation of $D^0$ is quite important, since it guides several components of the fusion methodology.

\begin{figure*}[t!]
\centering
\begin{tabular}{ccc}
\includegraphics[width=5.5cm]{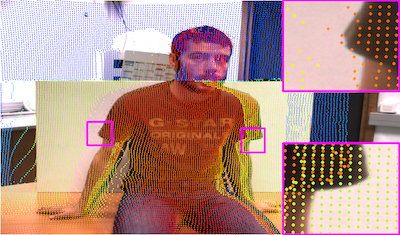} &
\includegraphics[width=5.5cm]{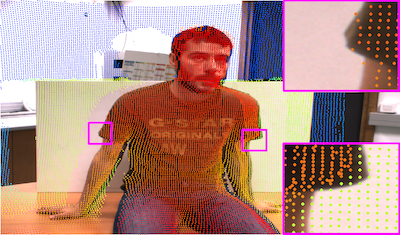} &
\includegraphics[width=5.5cm]{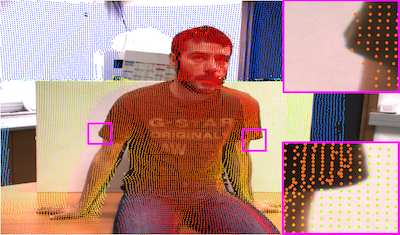} \\
(a) & (b) & (c)
\end{tabular}
\vskip -0.2cm
\caption{(\emph{Best viewed on screen}) (a) The mapping of depth data onto the left image causes artifacts in the presence of depth discontinuities. A cascade of (b) geometry-consistency and  (c) color-consistency filters refines the sparse disparity map. Depth values are color-coded from red (close) to blue (far). }
\label{fig:refinement}
\vskip -0.3cm
\end{figure*}

\section{Depth Initialization}\label{sec:depth_initialization}
A two-step approach is proposed in order to obtain the initial disparity map $D^0$. First, we refine \mbox{$D$$\downarrow$} to deal with mapping errors. Second, we upsample the refined sparse map in a novel way using color information to obtain the initial dense map $D^0$. As shown below, this leads to an initialization robust to depth discontinuities, which in turn helps the growing.

\subsection{Sparse-Depth Refinement}\label{sec:sparse_depth_refinement}
\begin{figure*}
\centering
\begin{tabular}{ccc}
\includegraphics[width=5.5cm]{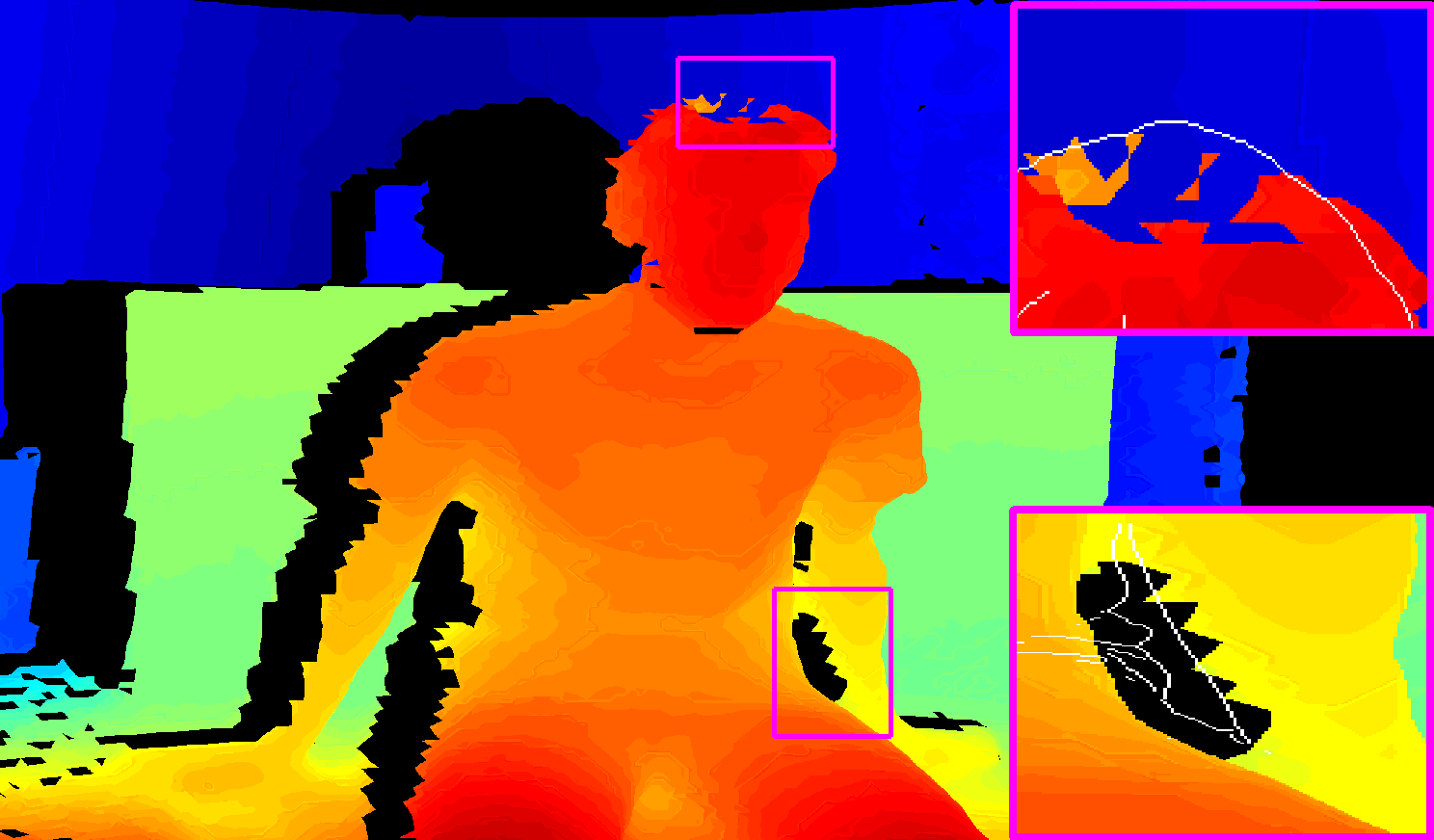} &
\includegraphics[width=5.5cm]{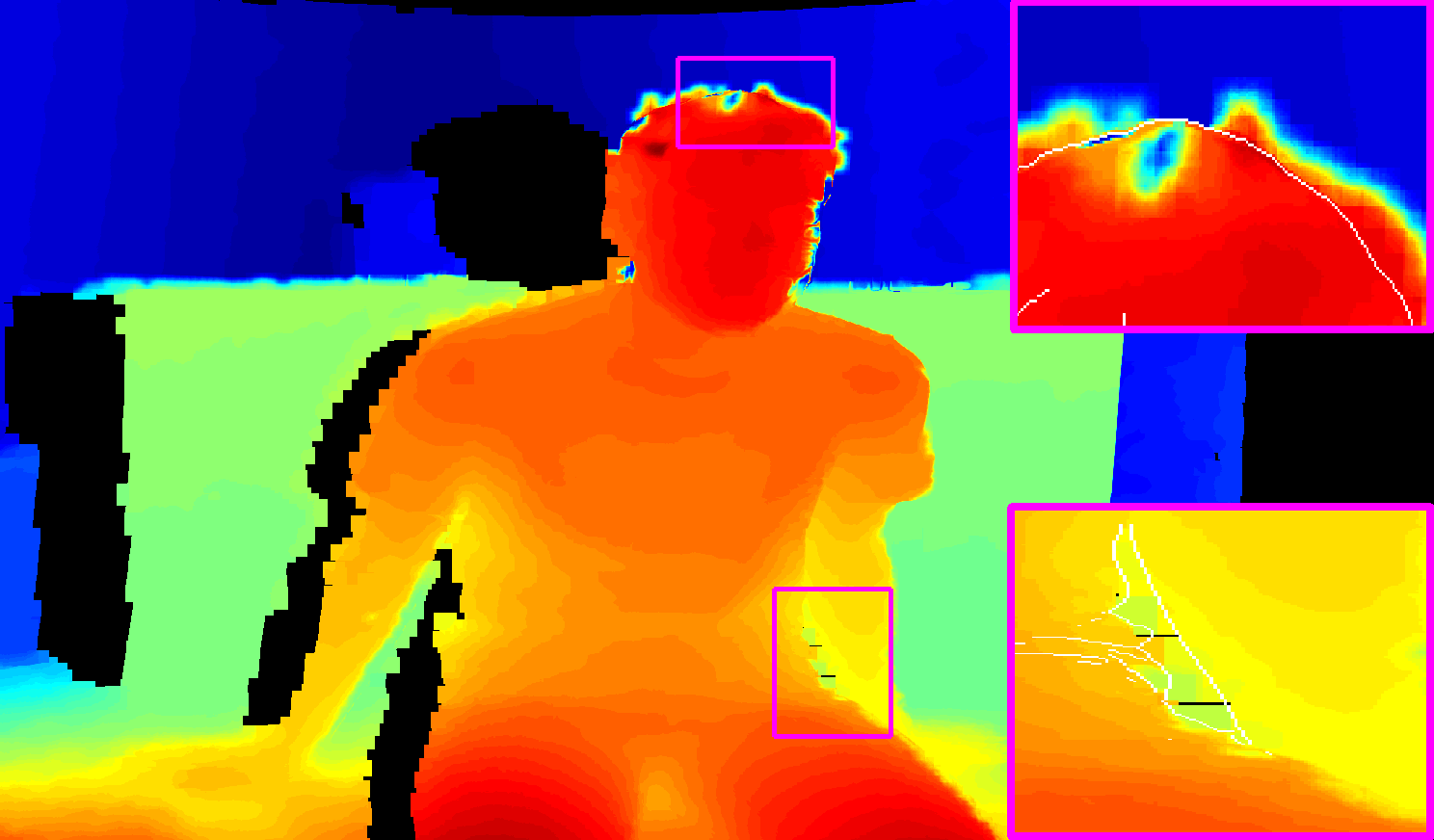} &
\includegraphics[width=5.5cm]{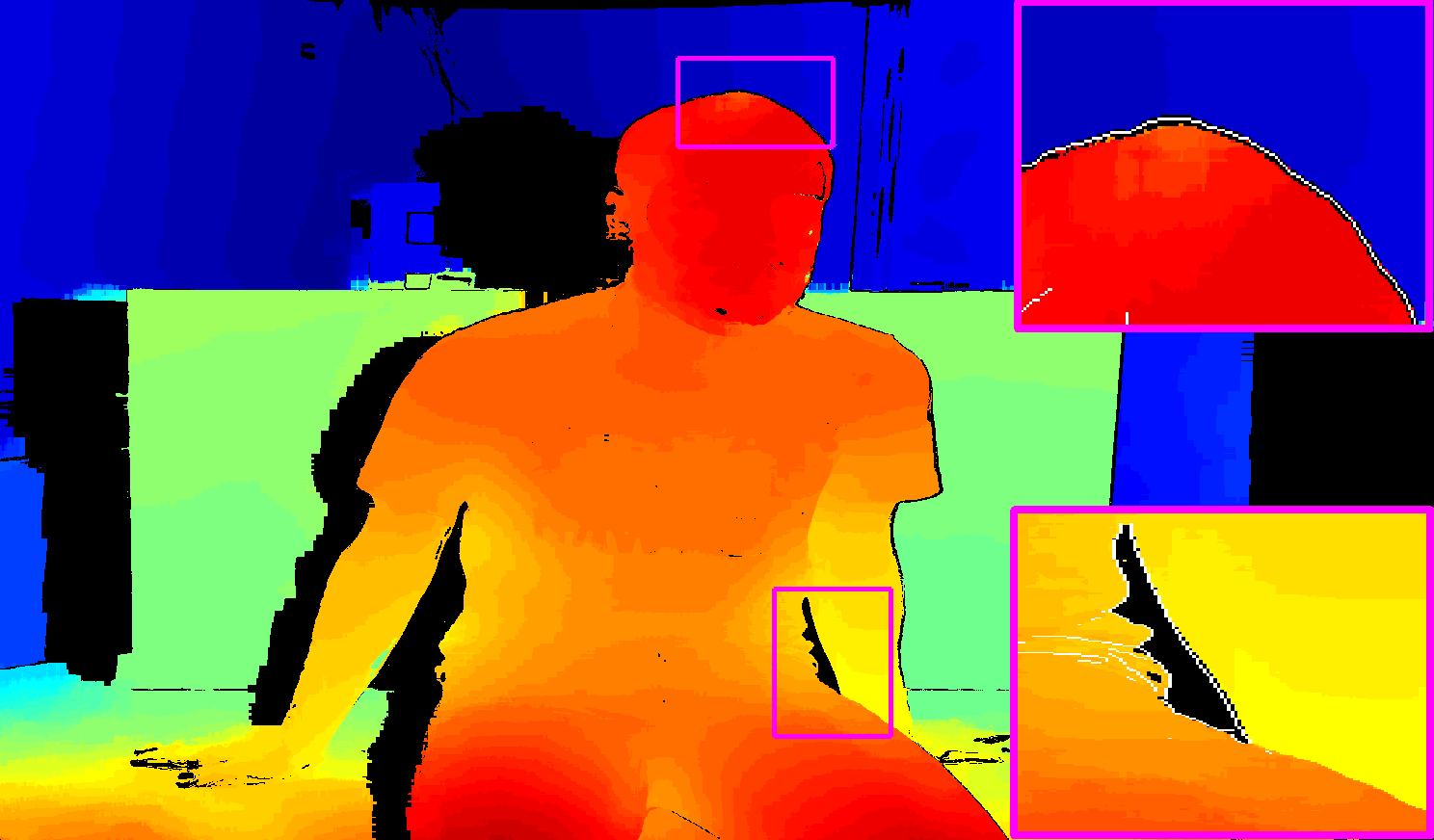}\\
(a) & (b) & (c) \\
\end{tabular}
\vskip -0.2cm
\caption{(\emph{Best viewed on screen}) Depth upsampling results using (a) triangulation-based interpolation~\cite{Gandhi12} after cutting big triangles,
(b) joint bilateral filter~\cite{Kopf2007jointBilateral} and (c) our method. The depth values are color-coded from red (close) to blue (far), while black areas correspond to non-available values. The white edges in close-ups show the color edges of the image.}
\label{fig:upsampling}
\vskip -0.25cm
\end{figure*}

We assume a camera setup with a depth camera mounted in between the two color cameras; other depth-stereo setups are discussed in~\cite{Nair2013}. Regardless of the depth sensor technology and type, the projection of the depth map onto the left and right images implies a parallax effect, and hence occlusions. Moreover, this causes gaps as well as areas with overlapping  depth data close to depth boundaries \cite{Gandhi12}. 
In the case of TOF cameras, these areas are further contaminated from \emph{jump-edge errors}~\cite{Foix11}, or `flying pixels'~\cite{Reynolds2011}, while a structured-light camera, e.g., Kinect, leaves more gaps due to the offset between the position of the light projector and the position of the infrared sensor. 
\mbox{Fig.~\ref{fig:refinement}(a)} illustrates the artifacts that we briefly discussed: flying pixels and depth-data overlap in the top and bottom closeups, respectively. In order to eliminate these artifacts we apply a geometry-consistency cascade of two filters: the first one removes isolated pixels (mostly flying pixels) and the second one keeps the foremost pixel inside a window to compensate for the above mentioned overlap. 
An example of applying this filtering is shown in Fig.~\ref{fig:refinement}(b). In practice this does not fully refine the sparse depth map. We still observe mismatches near depth discontinuities, because of depth bias and calibration errors. Therefore, a second filter that imposes color consistency is applied, as described below.

\begin{figure}[h!]
\centering
\includegraphics[width=5cm]{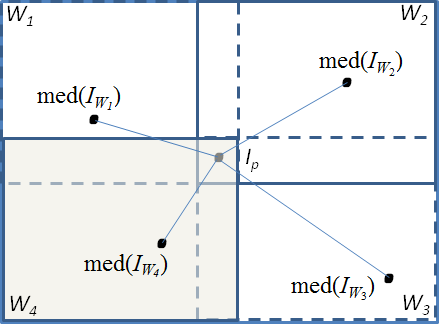}
\caption{The window split for the color-consistency filter. The pixel $p$ is linked with the closest (shaded) sub-window in terms of the color consistency (links represent color distances from $I_p$ to $\text{med}(I_{W_i})$).}
\label{fig:windows}
\end{figure}

We consider a window centered at \mbox{$p$$\downarrow$} and split into four equally sized sub-windows $W_i$, $i=\{1\hdots 4\}$, such that their intersection is only the pixel \mbox{$p$$\downarrow$} (see Fig.~\ref{fig:windows}). The output, $d_{p\downarrow}$, of the filter is:
\begin{equation}\label{eq:depth_refinement_filter}
  d_{p\downarrow} = \text{med}(D\!\!\downarrow_{W_{i^*}})
\end{equation}
with
\begin{equation}\label{eq:color_consistency}
  i^* = \arg\min_i \bigl(|I_{p\downarrow}-\text{med}(I_{W_i})|\bigr).
\end{equation}
The output $d_{p\downarrow}$ is the median disparity of the adjacent sub-window whose median intensity is closest to that at \mbox{$p$$\downarrow$}. For color images, the term $|I_{p\downarrow}-\text{med}(I_{W_i})|$ can be replaced by the average deviation from the median, over the color channels. This filter leads to a further refinement near depth discontinuities which are pathological areas for stereo algorithms. The result of this kind of filtering is shown in \mbox{Fig.~\ref{fig:refinement}(c)}. Note that both refinement filters apply to sparse locations only so that their complexity is negligible.

\subsection{Sparse-Depth Upsampling}\label{sec:depth_upsampling}


While a naive upsampling of the sparse disparity map could be performed, e.g., \cite{Gandhi12}, thereby producing an initial map, strong depth discontinuities are likely to contaminate such an interpolation. Alternatively, cross-bilateral filtering \cite{Petschnigg2004} or joint bilateral upsampling \cite{Kopf2007jointBilateral} may upsample the map using the HR color image as a guide. The filter support of these methods jointly counts on spatial and range (color) kernels. However, both methods apply a linear smoothing once the filter support per pixel is computed. Instead, we propose a non-linear upsampling strategy that builds on the median filter.


Specifically, the depth (or disparity) at a dense pixel location $p$ is initialized by 

\begin{equation}\label{eq:median_ipsampling}
d^0_p = \text{med}(D\!\!\downarrow_{\mathcal{N}_p^c})
\end{equation}
where $\mathcal{N}_p^c$ is a constrained neighborhood of $p$, that is $\mathcal{N}_p^c\subset\mathcal{N}_p$, which contains only sparse depth measurements whose color is consistent with $I_p$:
\begin{equation}\label{eq:sparse_neighborhood}
\mathcal{N}_p^c = \Bigl\{q\!\downarrow \,\, \big| \,\,
 g(I_p-I_{q\!\downarrow};\gamma_c)>\epsilon_c  \Bigr\}
\end{equation}
with $g(\xi;\gamma)$ being an exponential mapping (see Table~\ref{table:notations}).

Unlike common bilateral filters, our upsampling process makes a more definitive selection of pixels, thus preserving the depth discontinuities of the scene, while filtering out some of the noise in the depth data. Once the $\mathcal{N}_p^c$ is defined, one can optionally consider a spatial kernel and compute a weighted average instead, in order to better deal with slanted surfaces. Fig.~\ref{fig:upsampling} compares our initialization with the upsampling results obtained by~\cite{Gandhi12} and~\cite{Kopf2007jointBilateral}; in the detailed views, the intensity edges are also shown. The proposed method provides more discriminative depth boundaries that coincide with intensity edges. Missing values may be observed in highly textured areas, since $\mathcal{N}_p^c$ may be an empty set. In this example, the radius of $\mathcal{N}_p$ is $20$, $\gamma_c=10$, and $\epsilon_c=0.2$. The same radius is used for the method of~\cite{Kopf2007jointBilateral} while the scales for spatial and color kernel are $10$ and $20$ pixels, respectively. The geometry-consistency filter reasonably applies in all cases while our method benefits from our color-consistency refinement as well.

Since the median operator is chosen to account for outliers within a window, the mean operator can be invoked instead when the depth variance almost vanishes (homogeneous areas), thus drastically reducing the computational burden of the upsampling process. Since the vast majority of pixels belong to such areas, the complexity of our filter approaches that of joint upsampling filter~\cite{Kopf2007jointBilateral}. Note that the latter has been extended in~\cite{Mutto2012locallyConsistent} by integrating color segmentation results. 


\section{Depth-Stereo Fusion}\label{sec:fusion}

Let $d_p\in\mathcal{D}$ have $N$ possible discrete states; the goal is to estimate the disparity (state) of each HR image location through the MAP formulation (\ref{posterior_distribution}). Once $\mathcal{D}^0$ has been initialized, one can assume that $\mathcal S$ and $\mathcal{D}^0$ are conditionally independent, so that the posterior distribution of (\ref{posterior_distribution}) can be decomposed as
\begin{equation}\label{bayessian_model}
P(\mathcal{D}|\mathcal{S},\mathcal{D}^0) \propto
P(\mathcal{S}|\mathcal{D})\,
P(\mathcal{D}^0|\mathcal{D})\,
P(\mathcal{D}).
\end{equation}
As mentioned, global solutions are prohibitively expensive for high-resolution disparity maps. Therefore, we focus on approximate solutions that allow for the decomposition of the global optimization problem into many local (per-pixel) optimization problems. 

\begin{figure}
\centering
\includegraphics[width=7.0cm]{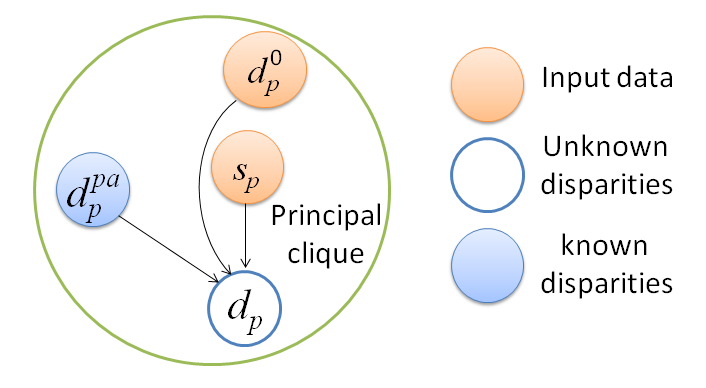}
\caption{The principal graph that is iteratively considered in our region-growing fusion method.}
\label{fig:principal_clique}
\end{figure}

The proposed method is based on the seeded region-growing framework~\cite{Adams1994,Cech07}, where the known message of a location (the parent) is propagated to its neighbor (the child). This implies that the estimation of $d_p$ is also conditioned by a parent \emph{known} disparity $d_p^{pa}$, hence dealing with the principal graph of \mbox{Fig.~\ref{fig:principal_clique}} for every pixel with unknown disparity (the visiting order of pixels is made clear later). As a result, if $P(d_p)$ is considered uniform, the posterior probability of $d_p$ can be written as\footnote{Strictly speaking, it is an approximation since $d_p^{pa}$ and $d_p^0$ may not be fully independent.}	
\begin{eqnarray}
P(d_{p}|s_p,d_p^0, d^{pa}_p) &=& P(d_{p}|d^{pa}_{p})P(d_{p}|s_p,d_p^0) \nonumber\\[.5ex]
&\propto& P(d_{p}|d^{pa}_{p})P(s_p|d_{p})P(d_p^0|d_{p}) \label{clique_equation}
\end{eqnarray}
where $d^{pa}_{p}$ is the parent of $d_{p}$ and the probability
\begin{equation}\label{prior}
P(d_{p}|d^{pa}_{p}) = \left\{
\begin{array}{l l}
    \frac{1}{2r+1} & \quad |d^{pa}_{p}-d_{p}|\leq r \\[.5ex]
    0 & \quad \text{otherwise} \\
  \end{array} \right.
\end{equation}
has a uniform distribution. In other words, the disparity range for each node is a function of the assigned disparity of his parent. We consider a narrow support area, i.e., a constant low value for $r$, e.g., $1$ or $2$; node-dependent parametrization of $r$ is left to future work.

\subsection{MAP as Energy Minimization}\label{sec:map_as_energy_min}
\begin{figure}
\centering
\includegraphics[width=8.0cm]{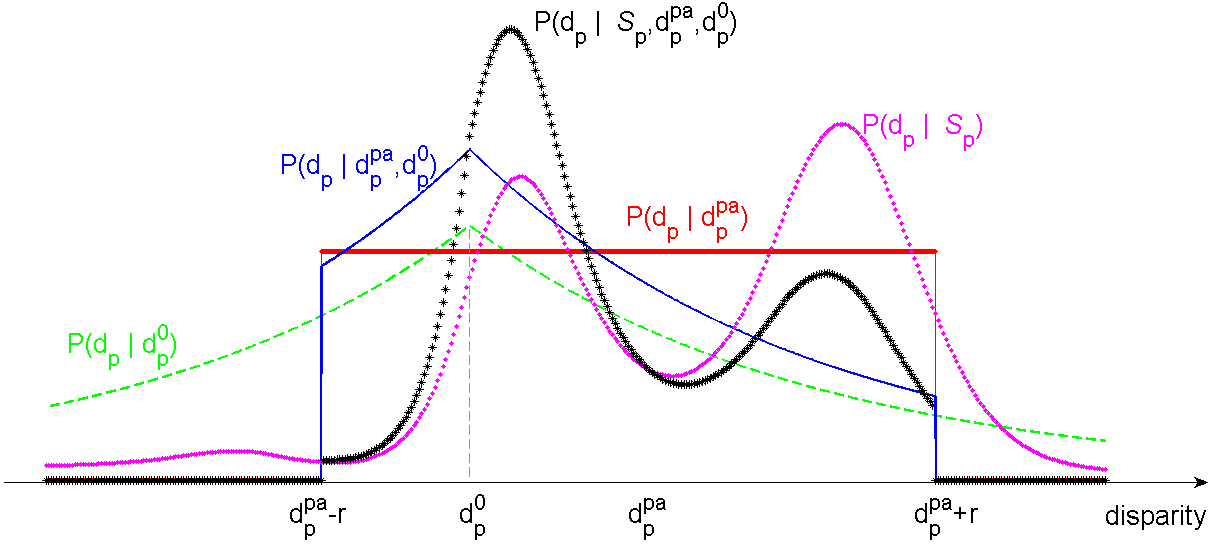}
\caption{An example of the probability distribution of $d_p$ constrained by different observations.}
\label{fig:probability_example}
\end{figure}

As is customary, likelihoods are chosen from the exponential family which leads to an energy minimization framework. Our model assumes an energy-dependent distribution (Boltzmann) for $P(s_p|d_{p})$ and a Laplacian one for $P(d_p^0|d_p)$:
\begin{eqnarray}
\label{similarity_likelihood}
P(s_p|d_{p}) &\propto & \exp\bigl(-E_S(d_{p}) \big/\lambda_S\bigr) \\[.5ex]
\label{regularization_term}
P(d_p^0|d_{p}) & \propto & \exp\bigl(-|d_{p}-d^0_{p}| \big /\lambda_D\bigr).
\end{eqnarray}
Based on (\ref{clique_equation}-\ref{regularization_term}), Fig.~\ref{fig:probability_example} shows an example with the distribution of $d_p$ being constrained by single or joint observations. Because of the exponential terms, the pixel-wise maximization of the posterior distribution reduces to the minimization of the \emph{local} energy
\begin{equation}\label{local_energy}
E(d_{p}) = E_S(d_{p})+E_D(d_{p})
\end{equation}
where $E_D(d_{p}) = \lambda|d_{p}-d^0_{p}|$ is the regularization term, $\lambda = \lambda_S/\lambda_D$ and $E_S(d_{p})$ is the (stereo) data-term which is defined below. The term $E_D(d_{p})$ guides the inference in textureless areas, while it penalizes mismatches due to depth discontinuities when the latter are well preserved in $D^0$. Notice that a smoothness constraint is implicitly enforced because of the prior term $P(d_{p}|d^{pa}_{p})$ owing to the low value of $r$. In other words, the support area of $E(d_p)$ is truncated as shown in Fig.~\ref{fig:probability_example}, and the pixel-wise optimization problem becomes
\begin{eqnarray}\label{local_energy_minimization}
&\min_{d_p} &E(d_p) \nonumber \\
&\text{subject to}~&|d_{p}-d^{pa}_{p}|\leq r.
\end{eqnarray}
The order of visiting pixels and solving (\ref{local_energy_minimization}) obeys a most-confident first-solved rule, as discussed in Sec~\ref{sec:proposed_algorithm}. Below, once the data-term is defined, we modify (\ref{local_energy}) to adaptively combine the data-term with the regularizer.

\subsection{The Data Term}
The proposed data-term benefits from two properties: its ability to provide energy for subpixel disparities and to adaptively consider pixel-wise similarities within a window. Apart from the advantage of a locally continuous map~\cite{Yang2007spatial}, it is more important to note that these properties lead to a more selective growing in areas with texture (owing to the former) and varying depth (owing to the latter) because of more accurate and reliable local energies. 

\subsubsection{Subpixel disparity}\label{sec:subpixel_disparity}
To obtain energy at interpixel locations, we make use of meta-disparities $(d_{p};t_{p})$ (see Table~\ref{table:notations}). The estimation of $t_p$ builds on~\cite{Psarakis05}, which has been shown to be superior to the parabola-fitting around the peak correlation: the pixel $p=(x,y)$ of the left image is matched with the subpixel $p'=(x+d_p+t_p,y)$ of the right image by defining a bounded correlation function $C_{d_{p}}(t_p)$ ($|C_{d_{p}}(t_p)|<1$) which is maximized with respect to $t_p$ for a given $d_p$. This allows us to define our data-term as
\begin{equation}\label{data_term1}
    E_S(d_{p};t_{p}^*) = 1-C_{d_{p}}(t_{p}^*),
\end{equation}
where
\begin{equation}\label{data_term2}
t_{p}^* = \arg\max_t\, C_{d_{p}} (t).
\end{equation}
Consequently, the total energy (\ref{local_energy}) is parameterized by $t_{p}^*$ as well. Note that if $t_p^*$ can be estimated via an analytic solution, the extra computational cost of subpixel disparity estimation is negligible. It becomes clear that, since $C_{d_p}(t_p^*)\geq C_{d_p}(0)$ in (\ref{data_term1}), more accurate energy values are estimated and more reliable messages are propagated. 

Since invariance to photometric distortions is important in stereo matching, we propose to adopt the normalized correlation coefficient, and one of its variants \cite{Moravec77}, for $C_{d_{p}}(t_p)$. While the former has been already extended to deal with subpixel disparities~\cite{Psarakis05}, the latter has never been extended before. Both use zero-mean vectorized forms of the windows around $p$ and $p'$, let $\mathbf{u}_L(x,y)$ and $\mathbf{u}_R(x+d+t,y)$, with the latter being written via a first-order Taylor approximation as $\mathbf{u}_R(x+d+t,y)\simeq \mathbf{u}_R(x+d,y) + t\Delta \mathbf{u}_R(x+d,y)$  where $\Delta$ is a difference operator along the $x-$axis. This is also the case in variational framework~\cite{Nair2012,Ruhl2012} where subpixel accuracy is obtained by the early interpolation of the intensity, rather than the late interpolation of correlation around the peak~\cite{Scharstein02}.

\subsubsection*{Enhanced Correlation Coefficient (ECC)}
The ECC function~\cite{Psarakis05} results from the integration of the above linear interpolation kernel into Pearson coefficient:
\begin{equation}\label{ECC}
    C_d^P(t) = \frac{\mathbf{u}_L^\top(\mathbf{u}_R+t\Delta \mathbf{u}_R )}{\|\mathbf{u}_L\|~\|\mathbf{u}_R+t\Delta \mathbf{u}_R\|}.
\end{equation}
If the denominator of (\ref{ECC}) is non-degenerate, then $C_d^P(t)$ is a quasi-concave function of $t$ and its maximization results in a closed-form solution~\cite{Psarakis05}.

\subsubsection*{Enhanced Moravec Correlation Coefficient (EMCC)}
The Moravec coefficient~\cite{Moravec77} replaces the denominator of (\ref{ECC}) with the mean of the variances. This also allows us to introduce a left-right symmetry, thereby estimating a left disparity $-t/2$ and a right disparity $t/2$ instead of $t$ (such a modification with ECC leads to a complex optimization problem). The \textit{enhanced Moravec correlation coefficient} (EMCC) is defined by
\begin{equation}\label{EMCC}
    C_d^M(t) = \frac{2(\mathbf{i}_L-t/2\Delta \mathbf{i}_L)^\top(\mathbf{i}_R+t/2\Delta \mathbf{i}_R )}{\|\mathbf{i}_L-t/2\Delta \mathbf{i}_L\|^2 + \|\mathbf{i}_R+t/2\Delta \mathbf{i}_R\|^2}.
\end{equation}
Note that one can easily show that the integration of an interpolation kernel into the cost function of~\cite{Gandhi12} is equivalent with the EMCC scheme.
Although (\ref{EMCC}) is a rational function of $t$, the next proposition guarantees that the maximizer has an analytic form. We refer the reader to the appendix for the proof and the exact maximizer.

{\proposition \it
A rational function of two second-degree polynomials, as in (\ref{EMCC}), attains at most one global maximum if the denominator is non-degenerate; its maximizer is given by a closed-form solution.
}

Note that, the estimation of $t$ could be unreliable for strictly homogeneous areas. The value of the window variance or entropy is a good criterion to assess the reliability of subpixel correction, and to enable it accordingly.

\subsubsection{Adaptive similarity aggregation}\label{sec:adaptive_aggregation}
The best-performing local stereo algorithms benefit from an adaptive cost aggregation strategy~\cite{Yoon06}. This strategy is based on the assumption that depth discontinuities are most likely to coincide with color discontinuities, so that each pixel within a window contributes differently to the (dis)similarity cost based on its spatial and color distance from the central pixel. However, only a few color edges correspond to depth edges and the above assumption should be followed only in the absence of any prior information about the depth. Since in our scenario the prior depth information is available, the spatial and color consistency can be replaced by a depth consistency term.

To be specific, we adopt here the exponential $g(\cdot)$ (see Table~\ref{table:notations}) to compute pixel-wise weights $w_q$:
\begin{equation}\label{eq:depth_consistency}
w_q = g\bigl(d^0_p-d^0_q;\gamma_d\bigr),
\end{equation}
with $q\in \mathcal{N}_p$. The weights apply element-wise to $\mathbf{u}_L$, $\mathbf{u}_R$, $\Delta \mathbf{u}_R$ and $\Delta \mathbf{u}_L$ in (\ref{ECC}) and (\ref{EMCC}). In other words, we compute the subpixel correction and the optimum local energy after down-weighting pixels in the window that belong to another surface compared to the one of the central pixel. It becomes clear now that not only the term $E_D(d_{p})$ but also the stereo term $E_S(d_{p})$ in (\ref{local_energy}) benefit from an upsampling method that is robust to depth discontinuities. Note that, even if the initial depth map is biased, it is sufficient enough to guide the aggregation step within the window.

\subsection{Adaptive Fusion}\label{sec:adaptive_fusion}
While a constant fusion may be reasonable for specific types of scenes (e.g.\ highly-textured scenes), an adaptive balance of the terms in (\ref{local_energy}) is usually preferred. This implies that the less we count on the data term $E_S(d_p)$, the more we should count on the regularization term $E_D(d_p)$ during the inference.\footnote{Here we omit the disparity correction $t$.} This suggests a convex combination when the scene point of $p$ is viewed by all cameras. 

A summary of methods that perform weighted fusion is discussed in~\cite{Nair2012}. However, most of them consider TOF-based weights for the regularizer (e.g.,~\cite{Zhu11}) which contradicts our goal of a sensor independent fusion. Moreover, directly using the confidence map of a TOF image is not a good strategy~\cite{Reynolds2011}. Therefore, we only rely on stereo data to obtain the mixing coefficients. It is well known in stereo or optical flow that the matching of a point is reliable when its associated image patch contains sufficient texture~\cite{Egnal2004,MacAodha2013}. Since a good indicator for the texture presence is the image entropy~\cite{Egnal2004}, an entropy filter provides us with am adequate reliability factor $e_p$ for each window centered at $p$. 

Let us now consider the left image as reference and compute the initial left-to-right disparities.  Likewise, we can build a right-to-left disparity map based on the right image, and a cross-checking of these maps can provide an estimation of the major occlusions due to strong depth discontinuities, with respect to the reference image. We refer to these areas as stereo-occlusions and we denote them as $\Omega_{SO}$. Recall now that some points in the left image are not seen by the depth camera, and that this gives rise to gaps in the initial disparity map which can be easily detected. We refer to these areas as depth-occlusions and we denote them as $\Omega_{DO}$. It becomes obvious that the evaluation of $E_S(d_{p})$ and $E_D(d_{p})$, in $\Omega_{SO}$ and $\Omega_{DO}$ respectively, should be avoided. Hence, we propose the following adaptive fusion
\begin{equation}\label{local_energy_adaptive}
E(d_{p}) = \eta_p^SE_S(d_{p})+ \eta_p^DE_D(d_{p})
\end{equation}
with the pair $(\eta_p^S,\eta_p^D)$ being defined as
\begin{equation}\label{eta_definition}
(\eta_p^S,\eta_p^D)\! = \!
\begin{cases}
(0, 1) & \text{if}~ p\in\Omega_{SO}\setminus\Omega_{DO} \\
(1, 0) & \text{if}~  p\in\Omega_{DO}\setminus\Omega_{SO} \\
(e_p, 1\!-\!e_p) & \text{if}~ p\in\Omega\setminus\Omega_{SO}\bigcup\Omega_{DO} \\
(inf, inf) & \text{if}~ p\in\Omega_{SO}\bigcap\Omega_{DO}
\end{cases}
\end{equation}
where $\Omega$ defines the whole image area and $e_p$ is the normalized output of the entropy filter. We intentionally add the last case in (\ref{eta_definition}) which shows that the fusion in $\Omega_{SO}\bigcap\Omega_{DO}$ is meaningless and a post-filling method should be followed.

\begin{figure}[b]
\centering
\includegraphics[height=2.0cm,width=2.8cm]{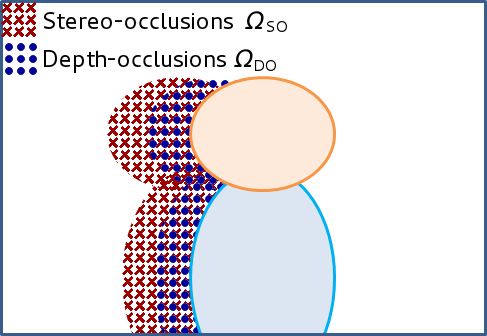}
\includegraphics[height=2.0cm,width=2.8cm]{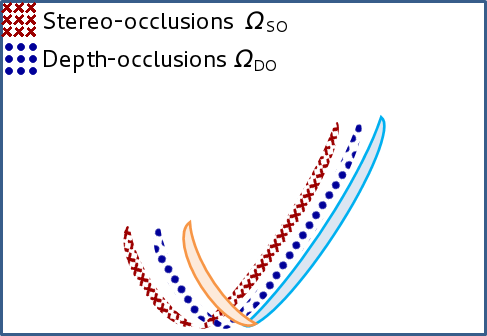}
\includegraphics[height=2.0cm,width=2.8cm]{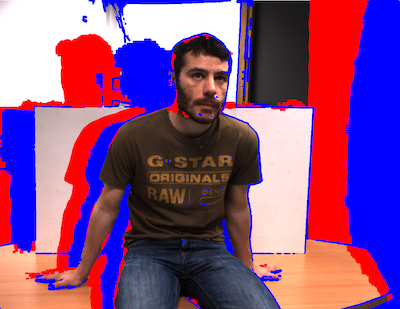}\\
\hskip 0.3cm(a)\hskip 2.4cm (b) \hskip 2.4cm (c)
\caption{(a) The validity of ordering constraint implies $\Omega_{DO}\subset\Omega_{SO}$ (depth camera is mounted between the color cameras); the shorter the baseline is, the more coincident these regions are. (b) $\Omega_{DO}$ and $\Omega_{SO}$ do not overlap when the scene contains very thin foreground objects and the ordering constraint is invalid. In (c), an estimation of $\Omega_{SO}$, $\Omega_{DO}$ for the example of Fig~\ref{fig:refinement} is shown.}
\label{fig:stereo_tof_occlusions}
\end{figure}


It is important to note that the so-called \emph{ordering constraint} in stereo is valid when large foreground objects appear in the scene, while it is violated when very thin objects are close to the camera (see Fig.~\ref{fig:stereo_tof_occlusions}). The former implies $\Omega_{DO}\bigcap\Omega_{SO}=\Omega_{DO}$ and the latter implies $\Omega_{SO}\bigcap\Omega_{DO}=\emptyset$. While the area $\Omega_{SO}\setminus\Omega_{DO}$ can always be predetected, the area $\Omega_{DO}\setminus\Omega_{SO}$ is safely predetected only when the ordering constraint is not valid.\footnote{It would be possible to detect all areas if one knows a priori that the constraint is \emph{everywhere} valid or invalid.} This is because $\Omega_{SO}$ is detected from the cross-checking of disparity maps that already suffer from depth-occlusions, since they are computed from the depth-to-stereo mapping. Ideally, if the complexity is not an issue, a stereo-occlusion detection scheme (e.g.~\cite{Sun05}) based on stereo image pair could be enabled beforehand. Therefore, we prefer to not grow the area of depth-occlusions. Optionally, these areas can be filled in a post-processing step. \mbox{Fig.~\ref{fig:stereo_tof_occlusions}(c)} shows the areas for the example of Fig~\ref{fig:refinement}. Points that have been removed during the refinement step (Sec.~\ref{sec:sparse_depth_refinement}) are also marked as depth-occlusions, while outliers in the initial maps can produce false positives for $\Omega_{SO}$, e.g. the red points on the right arm of the person. Our region-growing method, however, is able to compensate for such errors.

\subsection{The region-growing algorithm}\label{sec:proposed_algorithm}
As has been explained, our method solves pixel-based optimization problems in a region-growing manner, based on the seeds contained in \mbox{$\mathcal{D}$$\downarrow$}. Since the initial disparities may be noisy and biased, they do not reflect true matches as opposed to~\cite{Gandhi12}. This means that we exploit $d_{p\downarrow}$ to restrict the disparity range of its neighbor, but once a disparity value is assigned to the latter, \mbox{$p$$\downarrow$} is reset to a pixel $p$ with unknown disparity. For our convenience, the set \mbox{$\mathcal{D}$$\downarrow$} is augmented to an initial set $\mathcal{M}$ of meta-disparities $(d_p;0)$ with the same cardinality. We also denote with $\mathcal{N} (\mu_p)= \{\mu_p^j\}_{j=1}^4$ the image-based neighbors of $\mu_p$, that is, $\mu_p^j=(d_{\hat p}; t_{\hat p})$ with $\hat{p}$ being any of the four immediate neighbors of $p$. Note that $\mu_p^j$ does not necessarily belong to $\cal M$.



Algorithm~\ref{our_algorithm} describes the growing process. The algorithm starts by sorting the elements of $\cal M$ based on their energy value, while it initializes to false the visit- and assignment-flag of all candidate meta-disparities. Next, it considers the lowest-energy $\mu_p$ with false visit-flag and switches this flag to true. Then, it assigns values at \emph{each} $\mu_p^j\in\mathcal{N}(\mu_p)$ with false assignment-flag based on the following minimization scheme
\begin{equation}\label{local_minimization}
    (d^*_{\hat p};t^*_{\hat p}) =
\underset{t_{\hat p},\,|d_{\hat p}-d^*_{p}|\leq r}{\arg\min}
 E(d_{\hat p};t_{\hat p}),
\end{equation}
where
\begin{equation}\label{detailed_local_energy}
    E(d_{\hat p};t_{\hat p}) =
    \eta_{\hat p}^S E_S(d_{\hat p};t_{\hat p})  + \eta_{\hat p}^D E_D(d_{\hat p})
\end{equation}
and $d^*_{p}$ is the optimum disparity of the parent node of $\hat{p}$. Note that (\ref{detailed_local_energy}) extends (\ref{local_energy}) by adding the subpixel disparity parameter into the stereo term, and making the fusion pixel-dependent. In other words, what we do for each neighbor $\hat{p}$ with false assignment-flag is the following. For each candidate integer value $d_{\hat p}\in[d_p^*-r,d_p^*+r]$ the optimum $t_{\hat p}^*$ that minimizes the term $E_S(d_{\hat p};t_{\hat p})$ is obtained based on (\ref{EMCC}) or (\ref{ECC}) and the local energy $E(d_p;t_p^*)$ is computed from (\ref{detailed_local_energy}). Among the $2r+1$ values, the disparity minimizer $d_{\hat p}^*$ is finally chosen and the corresponding subpixel correction is assigned, as it is shown in (\ref{local_minimization}). Recall that $r$ has a low value in contrast to conventional stereo algorithms where it equals the whole disparity range. If $E(d_{\hat p}^*;t_{\hat p}^*)<T$, then the meta-disparity $(d_{\hat p}^*;t_{\hat p}^*)$ is pushed into $\mathcal{M}$ with respect to the sorting, and its assignment-flag becomes true. The above process is repeated with unvisited meta-disparities until the cardinality of $\cal M$ remains fixed. The visiting order depends on local energies, since the lowest-energy meta-disparity is always picked from the stack. The final set $\cal M$ corresponds to the final dense disparity map while a post-filling method can deal with missing disparities; their number heavily depends on threshold $T$. It is important to note that the algorithm cannot get stuck in a loop, because it propagates disparities in a tree structure, and a true visit-flag can never be reset to false. 

\begin{algorithm}[t!]\nonumber
\algsetup{linenosize=\scriptsize}
\caption{Stereo-Depth Fusion}
\begin{algorithmic}[1]
\REQUIRE Image-pair~$I_L, I_R$, set~$\mathcal{D}\!\downarrow$, Threshold~$T$.
\STATE Transform $\mathcal{D}\!\downarrow$ into a set $\mathcal{M}$ of meta-disparities with $t_p\!\!=\!0$.
\STATE Compute the initial disparity map $D^0$. 
\label{s:initial_map}
\STATE Sort $\cal M$'s elements based on their energy $E(\mu)$.\label{s:sorting}
\STATE Set both the visit and assignment flags $f_v(\mu)$, $f_a(\mu)$ to false, for all candidate meta-disparities, including $\mathcal{M}$.
\REPEAT
\STATE Consider $\mu_p$ with minimum energy and false $f_v(\mu_p)$
\STATE Set $f_v(\mu_p)=\text{true}$
\FORALL{ $\mu_p^j \in \mathcal{N}(\mu_p)$ with $f_a(\mu_p^j)=\text{false}$}
\STATE  $\mu_p^{j*} = \arg\min E(\mu_p^j)~$ (Eq. \ref{local_minimization})
\IF{ $E(\mu_p^{j*})<T$}
\STATE Set $f_a(\mu_p^{j*})=\text{true}$
\STATE Push $\mu_p^{j*}$ in $\cal M$ w.r.t. sorting
\ENDIF
\ENDFOR
\UNTIL{$\text{card}(\cal M)$ is fixed}
\STATE Compute the dense disparity map $D$ from $\cal M$
\RETURN $D$.
\end{algorithmic}\label{our_algorithm}
\end{algorithm}\vskip -0.6cm


\subsection{Comparison with Other Inference Formulations}\label{sec:discussion_comparison}

State-of-the-art stereo or fusion methods adopt an MRF model that provides a straightforward way to integrate multiple sensor data. If we recall equation (\ref{bayessian_model}), the term $P(D)$ can be written as a Gibbs distribution whose energy is a sum of potential functions over maximal cliques~\cite{Chou1990}, hence a sum of pairwise potentials when a 4-pixel neighborhood is considered. This in turn offers a smoothness term in the global energy equation that leads to piecewise smooth disparity map. There exist exact solutions for such models, under very specific conditions~\cite{Szeliski2008}.
Their computational complexity, however, becomes more and more severe as the number of states increases, thus becoming prohibitively expensive in the case of HR-stereo. Therefore, one has to focus on feasible approximate inferences that decompose the global optimization problem into a series of local optimization problems. 

\begin{figure}[t!]
\centering
\includegraphics[width=7.0cm]{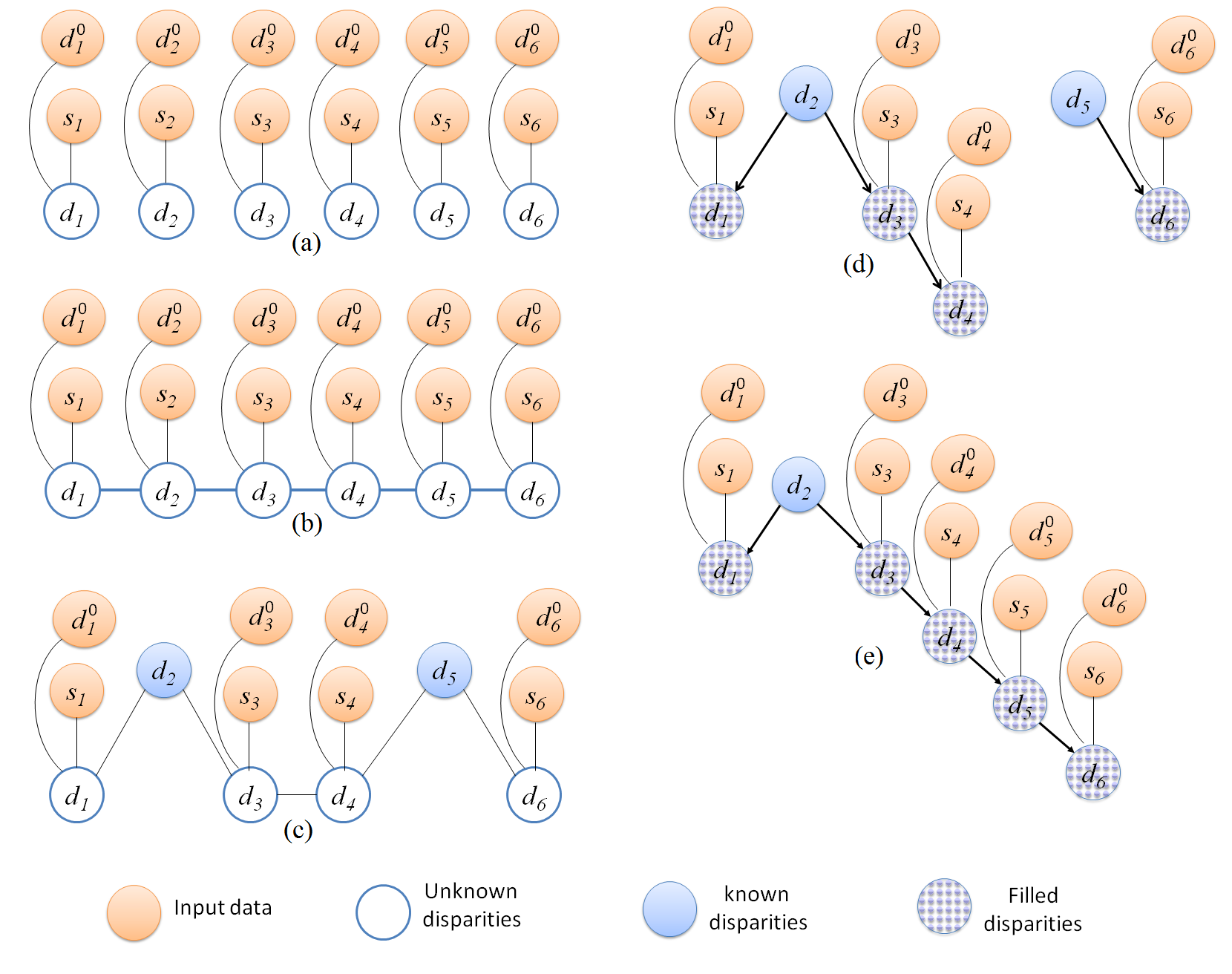}
\caption{Dependency network for (a) WTA and (b) MRF models in an 1D example. Initial and two final candidate graphs for the proposed scheme are shown in (c), (d) and (e) respectively. 
}
\label{fig:mrf_wta_ours}
\vskip -0.25cm
\end{figure}

MRF stereo generalizes the winner-take-all (WTA) approach~\cite{Scharstein02} which can be seen as the simplest inference. Fig.~\ref{fig:mrf_wta_ours}(a) and (b) show an 1D-grid example of the MRF and WTA networks. In essence, links between disparity nodes in WTA disappear and each node is connected only to input data (observations). This implies a uniform prior distribution $P(\mathcal{D})$ and the maximization of $P(\mathcal{D}|\mathcal{S},\mathcal{D}^0)$ reduces to a set of independent pixel-wise maximization problems of type $P(d_p|s_p, d_p^0)$. As with the proposed method, we tacitly assume, that ${s}_p$ are not intensities but they represent the stereo data of local windows centered at $p$ and its candidate correspondence. This is necessary in WTA since the likelihood cannot count on single pixels only, while it can optionally used in global MRF solutions as well. It is worth noticing that our algorithm switches to a WTA solution when $r=N$, since the visiting order of nodes and their connectivity becomes meaningless and the uniform distribution $P(d_p^{pa}|d_p)$ is not truncated anymore.

Another MRF approximation that finds pixel-wise optimizers without breaking the connectivity between nodes is the iterated conditional modes (ICM) method~\cite{Besag86}. After a proper initialization of all disparity nodes, ICM-like schemes visit one node (pixel) at a time and assign the disparity value that maximally contributes to the global posterior distribution. On the contrary, our approach requires the initialization of few nodes only, at least one in principle. Moreover, the graph in our case is a set of directed \emph{trees}, i.e., a forest, so that each non-root node has only one parent. It is important to note, however, that it is the output of our algorithm that defines the final network since nodes may be re-ordered, links may be cancelled and arrows may be reversed during the inference. 

To be more specific, let us consider the graph of Fig.~\ref{fig:mrf_wta_ours}(c) and let us assume the clamping of $d_2$ and $d_5$ to the observation $d_2^0$ and $d_5^0$. Starting from $d_2$, we can `propagate' $d_2^0$ to its neighbor $d_3$ (and $d_1$), i.e., to look for the optimum value of $d_3$ but being strictly conditioned by the value $d_2^0$. Note that ICM would look for the best assignment of $d_3$ value by taking into account both initial values $d_2^0$ and $d_4^0$. Moreover, ICM would search, in principle, among all (here $N$) states for the optimum $d_3$'s assignment, while our scheme looks only around $d_2^0$, namely $d_3$ takes values in $\{d_2^0-r,...,d_2^0+r\}$, with $r$ being a small integer. Once $d_3$'s assignment is done, $d_2^0$ can be passed to $d_1$ in a similar way. Next, another node is visited, here one of $d_3$ and $d_5$, and its disparity is propagated to its neighbors. As a result, the principal graph of Fig.~\ref{fig:principal_clique} is iteratively considered. Note that ICM assigns a new value to $d_3$ anyway, while our scheme invokes a criterion that validates the assignment. If $d_3$'s assignment is not valid, our algorithm will possibly assign a new disparity value to $d_3$ only after $d_4$'s assignment. Fig~\ref{fig:mrf_wta_ours}(d) and (e) show possible final graphs obtained by different realizations of our algorithm. In the example of Fig~\ref{fig:mrf_wta_ours}(e), the initial disparity of $d_5$ was cancelled by the validation process, thus all nodes were filled due to $d_2$.

It is now clear that, as opposed to WTA solution, the final inference we obtain depends on the visiting order of nodes. The visiting order of the ICM scheme is either fixed in advance (e.g.~raster-scanning), or random. Inspired by \cite{Chou1990} and \cite{Cech07}, however, we instead adopt a highest confidence first (HCF) scheme that suggests visiting the nodes based on their local evidence (energy). This means that we keep the nodes sorted with respect to their energy, and we visit each time the least-energy node that has not been visited yet. For instance, in Fig.~\ref{fig:mrf_wta_ours}(d), starting with $d_2$ implies that $d_2$ is more confident than $d_5$. The above validation process relies on thresholding the local energy as explained in Sec.~\ref{sec:proposed_algorithm}.
Table~\ref{table:inference_algorithms} summarizes some properties of MRF-based solutions (Graph-cut~\cite{Boykov98} and ICM~\cite{Besag86}), WTA, and our approach. 

\begin{table*}[t!]
\caption{Properties of inference algorithms in stereo and/or depth-stereo fusion.}
\begin{center}
\begin{tabular}{l|cccc}
  \hline
   & MRF (graph cuts) & MRF (ICM) & WTA & Proposed  \\
   \hline\hline
    Dependency network & undirected graph (MRF) & undirected graph (MRF) & independent minor graphs & independent directed trees (forest) \\
  Inference & exact* & approximate & approximate & approximate \\
  Prior distribution & Gibbs & Gibbs & uniform & truncated uniform  \\
  Invariance to visiting order & yes* & no & yes & no \\
  Disparity search range & full ($r=N$) & full ($r=N$) & full ($r=N$) & narrow ($r\ll N$) \\
  Complexity in HR stereo/fusion & too high & high & high & low \\ \hline
\end{tabular}\\
\end{center}
\vskip -0.2cm
*under specific conditions~\cite{Boykov98,Szeliski2008}
\vskip -0.2cm
\label{table:inference_algorithms}
\end{table*}



\section{Experiments}\label{sec:experiments}

In this section, we evaluate our algorithm, and quantitatively compare it to the state-of-the-art, based on both simulated and real data-sets. We also test our algorithm and provide qualitative comparison on a new and challenging dataset.

\subsection{Simulated Data}\label{sec:simu_results}
We use the Middlebury database, and focus on a challenging data-set which contains $1.5$MP images ($1300\times 1100$) along with ground-truth maps (GTM)~\cite{ScharsteinPal2007}. To simulate an LR disparity map from another viewpoint, we proceed as follows. Given the calibration parameters, we transform the ground-truth disparities into 3D points, as viewed from the midpoint of the baseline. We then apply a 3D rotation to the point-set and we downsample the points by a factor of $10$. The rotation is such that the average disparity bias is more than $2$ pixels. Finally, we back-translate the points into sparse biased disparities and we add colored noise, that is, a 2D mid-frequency sinusoidal signal with peak-to-peak distance equal to $2\sigma$, where $\sigma$ denotes the noise deviation. Note that \cite{Kopf2007jointBilateral,Yang2007spatial,Gandhi12} only downsample the GTMs in their experimental setup. Algorithm performance is quantified in terms of the so-called \emph{bad matching pixels} (BMP) percentage in the non-occluded areas~\cite{Scharstein02}, i.e., $(1/N_o)\sum_{p}(|D_p-G_p|>\delta)$ where $G$ is the GTM and $N_o$ is the number of the non-occluded pixels. While the threshold level $\delta = 1$ is mostly used for mid-resolution images, HR stereo justifies the value $\delta=2$ as well~\cite{Geiger10}; a value $\delta<1$ is chosen when subpixel accuracy is to be evaluated.

The same parameter settings are used for our method, in all of the experiments. The radius of the upsampling filter is $20$ and the values $\gamma_c$ and $e_c$ are $10$ and $0.2$ respectively. Because of the propagation strategy, we choose a relatively small window, i.e., $9\times9$. The local energy in (\ref{local_energy}) and the weights in (\ref{eq:depth_consistency}) are obtained with $\lambda=0.01$ and $\gamma_d=5$, respectively. We enable subpixel correction when the normalized entropy in the (left) window is above $0.4$. As mentioned, we use a fixed (and strict) search range around the disparity parent, that is $r=1$, which leads to the most efficient solution. As for the threshold, we set $T=0.5$. Note that the energy validation threshold implies a tradeoff between accuracy and density. We recommend setting a middle threshold value and post-filling sparse missing disparities, e.g., with the upsampling filter, rather than using a high threshold that incorporates erroneous disparities in a fully dense map. The density obtained with this strategy is about $90\%$ in HR images. As with all algorithms, large remaining gaps are filled with a streak-based filling~\cite{Geiger10}. We refer to our methods as Fusion-ECC (F-ECC) and Fusion-EMCC (F-EMCC). 

Before comparing with the state-of-the-art, we show the performance gain in terms of the new modules that are integrated compared to EPC method~\cite{Gandhi12}, thus quantifying the contribution of the new data-term and the adaptive fusion. Note that~\cite{Gandhi12} (EPC) follows a seed-growing approach by using a quadratic model for both stereo and depth consistency terms respectively. To better evaluate the contribution of the enhanced correlation coefficients presented in the data-terms, we use LR images ($450\times 375$ on an average) whose GTM contains subpixel disparities. The noise deviation in the sparse map is $\sigma=2$. We intentionally do not fill any remaining large gaps, in order to assess the net contribution of each module, and we compute the error for the filled area only ($85\%$ density). Table~\ref{table:modules} shows the BMP error averaged over eight images. We also evaluate our approach when none of the modules are enabled, i.e. pure correlation is used along with a fixed fusion of the terms; we refer to this method as \emph{simple fusion}. All of the variants start from the same initial depth map, obtained by our upsampling method. As can be seen, both the data-term and the adaptive fusion process contribute to a better reconstruction, compared to  simple fusion. The use of the proposed energy data-term leads to more accurate results, while the adaptive fusion eliminates large errors (see the error with $\delta=2$). Even the simple fusion has a lower BMP than EPC, owing to the different stereo- and depth-consistency terms (we use a linear model for the latter). The two proposed criteria behave similarly, with the F-ECC being slightly more accurate, since it achieves higher correlation values (see also Table~\ref{table:HR_1}). Note that the results are systematically worse if we use the initialization of~\cite{Gandhi12}. 


\begin{table}[b!]
\centering
\caption{Contribution of various modules of the proposed algorithm (BMP error averaged over eight LR images).}
\begin{tabular}{|p{1.65cm}|c|c|c|}
  \hline
  & \multicolumn{3}{c|}{BMP ($\%$) for $\delta=0.5$ / $\delta=1$ / $\delta=2$} \\ \hline
  & F-ECC & F-EMCC & EPC~\cite{Gandhi12} \\ \hline
  Simple Fusion 				& $22.4$ / $7.7$ / $3.4$ & $23.3$ / $7.9$ / $3.5$ & \multirow{3}{*}{$33.0$ / $11.6$ / $3.7$} \\ \cline{1-3}
  Data-term  					& $16.8$ / $6.4$ / $3.2$ & $17.2$ / $6.7$ / $3.3$ & \\ \cline{1-3}
  Data-term+ Adap. fusion 	& $14.6$ / $5.8$ / $2.4$ & $15.0$ / $6.2$ / $2.7$ & \\ \hline
\end{tabular}
\label{table:modules}
\end{table}

To compare with the state-of-the-art, we implemented the upsampling methods of \cite{Kopf2007jointBilateral} and \cite{Yang2007spatial} (two-view version), as well as the MRF-based fusion~\cite{Zhu08} by using the MRF-stereo toolbox of \cite{Szeliski2008}, referred here to as F-MRF. While \cite{Zhu08} uses belief propagation, we experimentally found that Graph-Cuts~\cite{Boykov98} perform better. Specifically, we tried ten different parameter settings and we found that the best performing algorithm is the expansion mode with Birchfield-Tomasi cost~\cite{Birchfield98} truncated at $7$, linear disparity differences truncated at $5$ and quadratic cost for the smoothness-energy; the weights for the stereo-, depth- and smoothness terms were $1$, $1.2$ and $10$. We refer here to~\cite{Zhu08} instead of~\cite{Zhu11} for MRF-based fusion since the latter relies on a TOF-based reliability fusion which cannot be implemented here. 

\begin{figure}[b!]

\includegraphics[height=28mm, width=36mm]{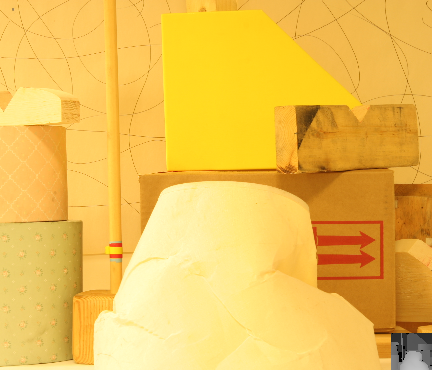}
\includegraphics[height=28mm, width=48mm]{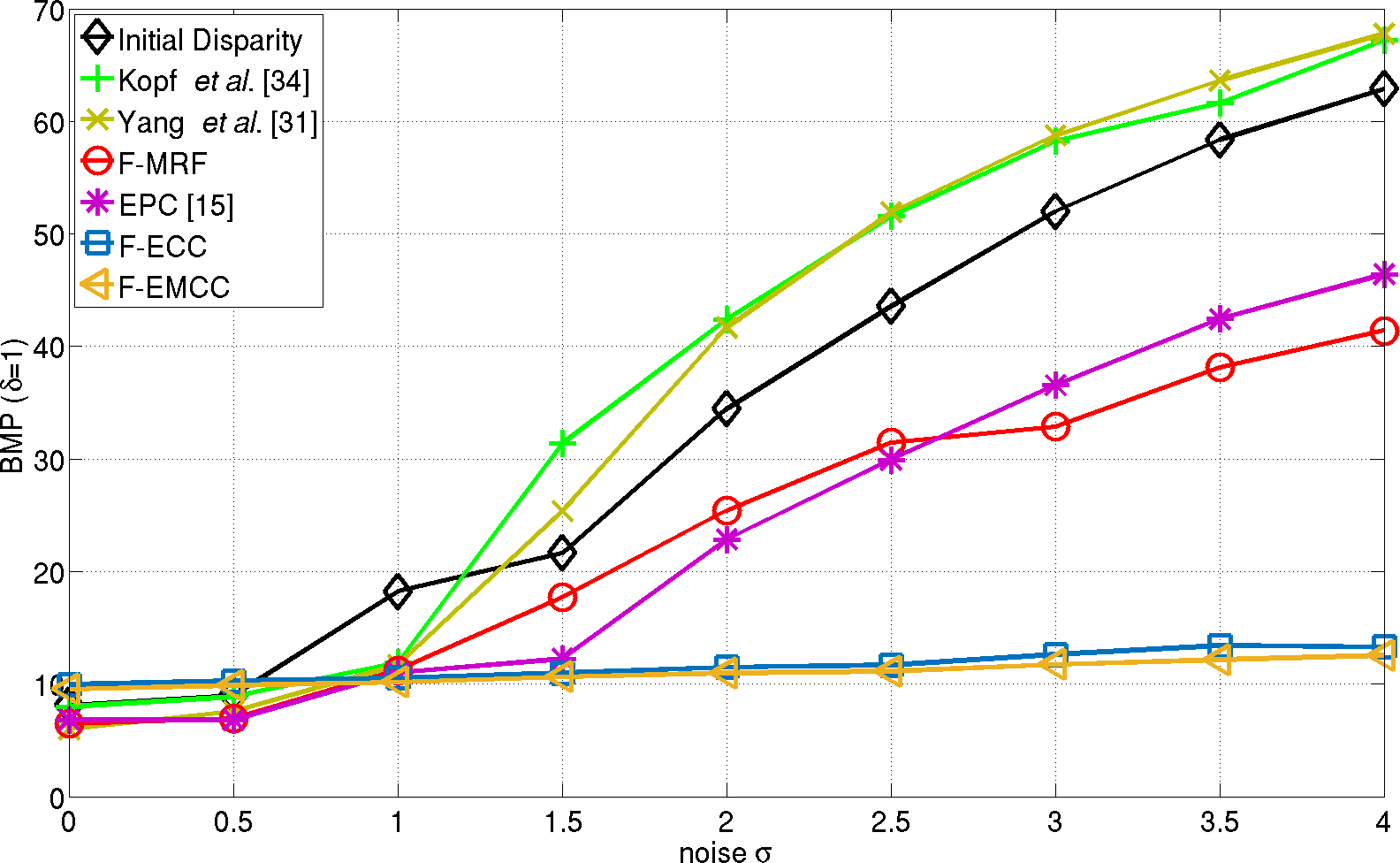}
\caption{Left image with superimposed depth map (left) and BMP curves (right) for the
HR stereo pair \emph{Lampshade1}.}
\label{fig:BMP_noise}
\vskip -0.3cm
\end{figure}

Fig.~\ref{fig:BMP_noise} plots the BMP curves of the upsampling and fusion algorithms as a function of the noise deviation for the challenging low-texture HR image \emph{Lampshade1}. We just add noise here in the down-sampled GTMs. Except for~\cite{Kopf2007jointBilateral}, all schemes start from the same HR map, obtained by a \emph{naive interpolation}, while its BMP curve is plotted as well. As can be seen, F-MRF and EPC are more affected by the noisy prior disparity, in contrast to the proposed algorithm, which is less sensitive to initialization. It is clear that the pure up-sampling methods provide acceptable results only when the initial LR disparity map is very accurate.

\begin{table*}
\centering
\caption{BMP for \emph{high-resolution} disparity maps of the Middlebury dataset with $\delta=1$.}
\footnotesize
\begin{tabular}{|l|l|c|c|c|c|c|c|c|c||c|}
  \hline
 &   & Lampshade1 & Art & Monopoly & Rocks2 & Reindeer & Bowling2 & Baby1 & Moebius & Average \\
  \hline
\multirow{6}{*}{Stereo} &
   GCS~\cite{Cech07} 								& 25.1 & 20.0 & 52.4 & 4.5  & 14.1 & 25.4 & 18.1 & 19.5 & 22.4\\
&  ELAS~\cite{Geiger10} 								& 15.2 & \underline{12.2} & 38.1 & \underline{2.6}  & \bf{5.6} & 14.6 & 10.4 & 14.0 & 14.1 \\
&  GraphCuts~\cite{Kolmogorov01} 						& 32.3 & 26.3 & 62.1 & 10.8 & 28.8 & 43.1 & 15.6 & 19.5 & 29.8\\
&  CSBP~\cite{Yang2010ConstantSpaceBP} 				& 39.4 & 26.6 & 61.8 & 8.3 & 20.3 & 31.9 & 16.4 & 20.0 & 28.1\\
&  FastAgg~\cite{Rhemann2011fastCost} 					& 26.8 & \bf{10.8} & 44.6 & \bf{2.0} & \underline{8.1} & 15.6 & 11.5 & 14.9 & 16.8\\
&  NonLocalAgg~\cite{Yang2012_nonLocaLCostAggregation} 	& 25.0 & 23.5 & 32.6 & 10.7 & 24.3 & 28.8 & 18.0 & 17.2 & 22.5\\
\hline
\multirow{2}{*}{Upsamping} &
  Yang \emph{et al.}~\cite{Yang2007spatial}        			& 46.9 & 47.4 & 48.5 & 41.8 & 42.9 & 46.3 & 36.9 & 41.9 & 44.1\\
&  Kopf \emph{et al.}~\cite{Kopf2007jointBilateral} 			& 43.7 & 46.5 & 45.4 & 39.4 & 40.9 & 42.1 & 35.1 & 41.8 & 41.8\\
 \hline
 \multirow{5}{*}{Fusion} &
   EPC~\cite{Gandhi12}  								& 20.7 & 17.8 & 20.7 & 4.2  & 11.8 & 20.8 & 11.9 & 12.0 & 14.9\\ 
 &  F-MRF (GC)~\cite{Zhu08}  							& 16.3 & 17.1 & 25.7 & 4.5 & 12.1 & 20.0 & 10.1 &  \underline{10.5} & 14.5\\ 
&  F-MRF (ICM)  										& 23.1 & 27.9 & 43.4 & 14.7 & 20.1 & 24.8 & 17.9 & 20.1 & 24.0\\
&  F-ECC 											& \underline{8.4} & 14.0 & \underline{7.6} & 2.9 & 8.7 & \bf{8.5} & \underline{4.4} & \bf{9.0} & \bf{7.9}\\ 
&  F-EMCC 										& \bf{8.2} & 14.9 & \bf{7.5} & 3.1 & 9.0 & \bf{8.5} & \bf{4.2} & 10.8 & \underline{8.2}\\ 


   \hline


\end{tabular}
\label{table:HR_1}
\vskip -0.25cm
\end{table*}

We now proceed with a detailed comparison including efficient and well-known stereo algorithms as well. Specifically, we include four recently proposed methods, \cite{Geiger10} (ELAS), \cite{Cech07} (CGS), \cite{Rhemann2011fastCost} (FastAgg), \cite{Yang2012_nonLocaLCostAggregation} (NonLocalAgg) and two MRF-based stereo algorithms, Graph Cuts (GC)  \cite{Kolmogorov01} and constant-space belief propagation (CSBP)~\cite{Yang2010ConstantSpaceBP}. The top-performing local algorithms, FastAgg and NonLocalAgg, build cost volumes that depend on both the image size and the disparity range. This leads to a huge memory footprint ($\sim3$GB) in the case of HR images, and the authors' implementations could not be run as is. In order to be able to run FastAgg, the cost volume has been split into slices and cached on disk. The NonLocalAgg was run using the maximum allowed resolution while the disparity map produced by the algorithm was finally upsampled. Note that both methods invoke a left-right consistency checking, combine color and gradient information and enable refinement steps. Authors' implementations for ELAS, GCS, GraphCut, CSBP (local-minima version$+$bilateral post-processing) were used in the comparisons, with the default settings suggested by the authors. We also implemented the ICM algorithm for the MRF-based fusion using the same parameters with GC. For a fair comparison, all fusion schemes merge stereo data with the \emph{same} initial disparity map, which is obtained here by our upsampling process. Due to the simulated experimental setup, however, the error of the initial map obtained from this process is very close to that of Kopf \emph{et al.}'s method  \cite{Kopf2007jointBilateral} (the average difference of their BMP error is below $0.5$) and is thus omitted.



Table \ref{table:HR_1} provides the BMP error for eight HR images with error threshold $\delta=1$, while the corresponding table for $\delta=2$ is given in the appendix.  
Bold and underlined numbers mark the lowest and second lowest errors per column. Weakly textured scenes (\emph{Lampshade1}, \emph{Monopoly}) seem to be problematic for conventional stereo algorithms, while cylindrical surfaces (\emph{Bowling2}, \emph{Baby1}) present another challenge. It is not surprising, however, that stereo methods outperform fusion methods in highly textured images (e.g. \emph{Rocks2}), or in images with many thin objects (e.g. \emph{Art}), since the sparse noisy initial map negatively affects the fusion. ELAS and FastAgg behave better than other stereo algorithms. Similar results would be expected from the NonLocalAgg method, if we were able to run it at full resolution.


Stereoscopic and depth data are better fused in general by the proposed criteria than the other fusion methods. EPC and up-sampling methods verify the sensitiveness to their initialization, with the former being more effective due to the fusion process. 
As far as the MRF solution is concerned, the benefit due to the depth data is verified from the results, i.e. the F-MRF (GC) scheme behaves better than pure-stereo GC. Moreover, GC in fusion provides better results than the simple ICM algorithm. Unlike the proposed criteria, F-MRF deals better with thin objects, since it does not aggregate costs in a window, hence the lower error in \emph{Art} and \emph{Reindeer}. Recall that \cite{Zhu08} refines the TOF depth map based on stereo data, without increasing the resolution. By putting aside the high complexity, it seems that MRF-based solutions need to be reformulated for HR stereo-depth fusion, e.g. high-order connectivity might be more helpful, semi-global solutions could be investigated, and conditional random fields~\cite{ScharsteinPal2007} might need to be extended to the fusion framework.

Between the two proposed criteria in our fusion scheme, ECC and EMCC, it is the image content that makes one outperform the other. While F-ECC may be slightly better on average, F-EMCC deals better with images of very low texture, e.g. \emph{Monopoly}, which supports  Moravec's argument for introducing MCC~\cite{Moravec77}. Moreover, F-EMCC is more affected by the filling, as it provides less dense maps than F-ECC, provided that the threshold is the same.

\begin{figure}
\centering
\includegraphics[width=8.0cm]{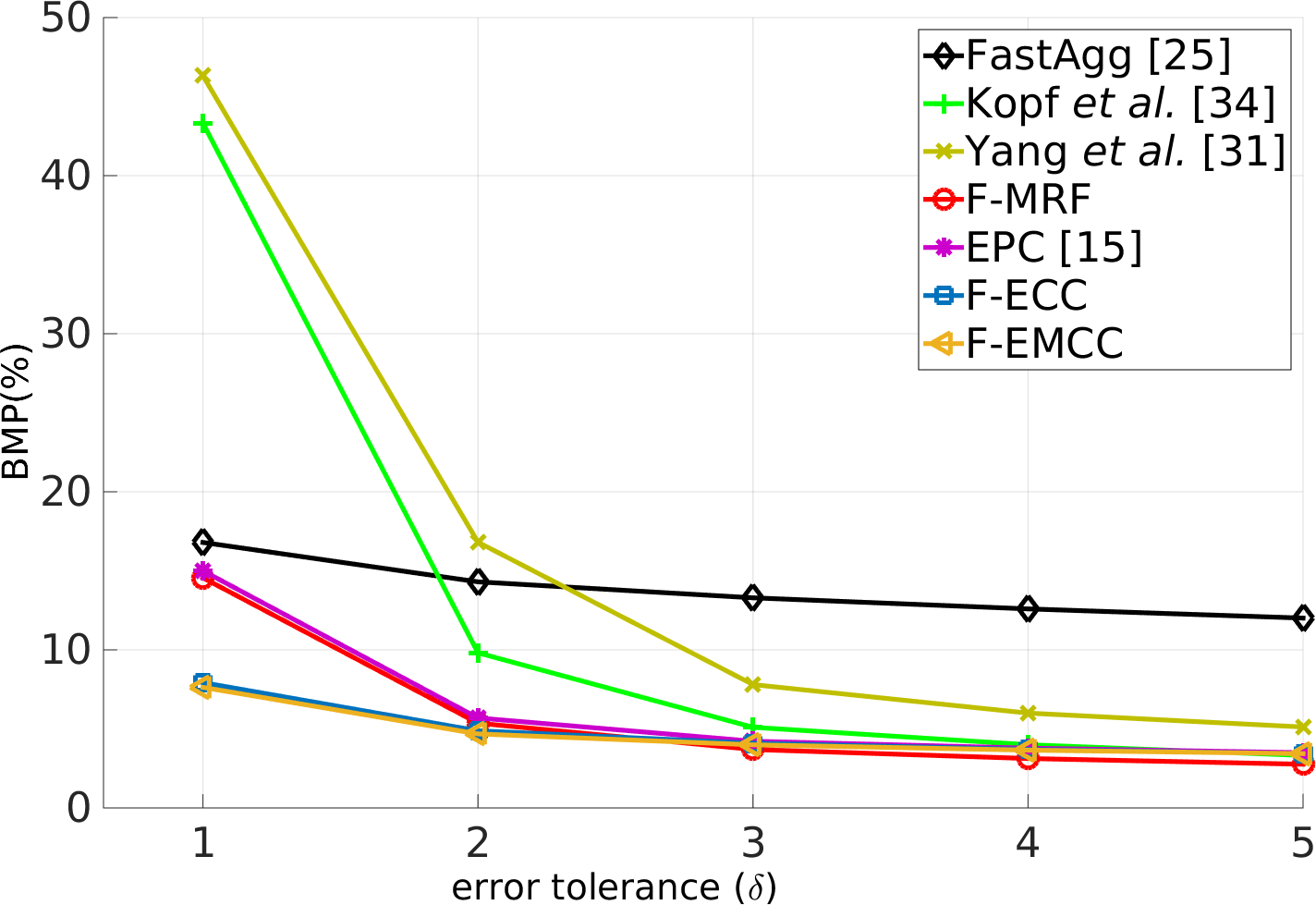}
\caption{The BMP error of fusion algorithms averaged over the eight test images as a function of the error tolerance $\delta$.}
\label{fig:average_BMP_scores}
\vskip -0.3cm
\end{figure}

Fig.~\ref{fig:average_BMP_scores} shows the average performance of all fusion competitors as a function of the error threshold $\delta$, while the stereo baseline of FastAgg is added for reference. In essence, this figure reflects the distribution of errors. Evidently, the contribution of the LR depth prior in the fusion schemes is verified, as opposed to the stereo baseline whose performance is bounded. The proposed schemes are more accurate compared to the fusion baselines. However, F-MRF provides lower errors when the tolerance is not that strict ($\delta>2$) owing to its global  smoothness constraint.
Note that the performance of~\cite{Kopf2007jointBilateral} approaches the performance of the fusion schemes as $\delta$ is increasing. The other upsampling method of~\cite{Yang2007spatial} seems to produce large errors.  


\begin{table}[b!]
\centering
\caption{Average execution times of algorithms for the HR Middlebury data-set~\cite{ScharsteinPal2007} in a $2.6$GHz machine.}
\footnotesize
\renewcommand{\arraystretch}{0.84}
\begin{tabular}{|m{2.4cm}|c|c|c|c|c|}
  \hline
  & Time(sec) & Matlab & C/C++ & GPU & SSE \\
  \hline
  GCS~\cite{Cech07} & $1.2$ & $\checkmark$ & $\checkmark$ &  & \\
  ELAS~\cite{Geiger10} & $1.0$ &   & $\checkmark$ &  & $\checkmark$ \\
  GC~\cite{Kolmogorov01} & $>10^3$ &   & $\checkmark$ &  &  \\
  CSBP~\cite{Yang2010ConstantSpaceBP} & $15$ &   & $\checkmark$ &  &  \\
  FastAgg~\cite{Rhemann2011fastCost} & $>10^3$ & $\checkmark$ &   &   &   \\
  NonLocalAgg~\cite{Yang2012_nonLocaLCostAggregation} & $5.0$ &  & $\checkmark$ &  &  \\
  \hline
  Yang et al.~\cite{Yang2007spatial} &  $>10^2$ & $\checkmark$ &  &  & \\
 Kopf et al~\cite{Kopf2007jointBilateral} & $88$ & $\checkmark$ &  &  & \\
 \hline
  EPC~\cite{Gandhi12} & $1.5$ & $\checkmark$ & $\checkmark$  &  & \\
 F-MRF (GC)~\cite{Zhu08} & $>10^3$ &  & $\checkmark$ &  &  \\
 F-MRF (ICM)~\cite{Zhu08} & $>10^2$ &  & $\checkmark$ &  &  \\ 
 Our upsampling & $95$ &  $\checkmark$ & &  &  \\
 F-ECC, F-EMCC & $2.0$ & $\checkmark$ & $\checkmark$ &   &   \\
\hline\hline
 Upsampling of~\cite{Gandhi12} & $<0.1*$ &  &$\checkmark$ & $\checkmark$ &  \\
 F-ECC, F-EMCC & $0.3*$ &   & $\checkmark$ &  $\checkmark$ & $\checkmark$  \\
   \hline
 \multicolumn{6}{l}{  \scriptsize * time needed for the 2MP images of our real data-set (see Fig.\ref{fig:real_scenes})}
\end{tabular}
\label{table:times}
\end{table}

Computational efficiency is an important feature of any depth-stereo fusion method. Table~\ref{table:times} shows the execution times of the algorithms for the simulated data, as well as some of their implementation details. A combined Matlab-C version of our fusion algorithm requires $2.0$s per image triplet while it takes $0.3$s (for 2MP images) when GPU hardware is used for some initializations, and the SSE instruction set accounts for the online computation of correlation values between windows. We also developed a GPU-based implementation of a triangulation-based upsampling (interpolation) that takes less than $50$ms. This allows one to envisage real-time execution of the proposed depth-stereo fusion framework, despite the high resolution. Note that it is the Matlab implementation that makes FastAgg and upsampling schemes slow, while the time for the NonLocalAgg method is based on matching at half resolution. We also point out that CSBP attains a solution in a reasonable time, despite its MRF-based formulation.



\subsection{Real Data}
\begin{figure}
\includegraphics[width=2.85cm]{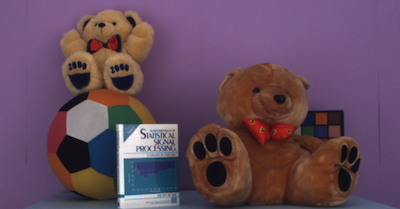}
\includegraphics[width=2.85cm]{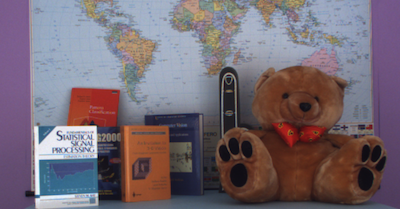}
\includegraphics[width=2.85cm]{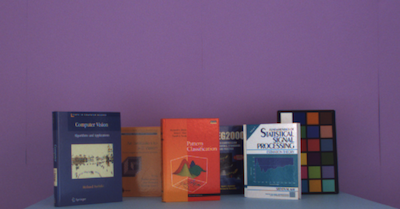}\\
\text{~~~~~~~~~~~}(A) \hspace{2cm} (B) \hspace{2cm} ~~(C)
\caption{The three scenes (cropped) of the dataset used in~\cite{Mutto2012locallyConsistent}.}
\label{fig:mutto_scenes}
\vskip -0.4cm
\end{figure}

\begin{table}[t!]
\centering
\caption{Evaluation on the dataset of Dal Mutto~\emph{et al.}~\cite{Mutto2012locallyConsistent}}
\begin{tabular}{|l|c|c|c||c|}
   \hline
    & \multicolumn{4}{c|}{MSE of disparity estimation} \\  \hline
  & Scene A & Scene B & Scene C & Average \\ \hline
  Stereo~\cite{Rhemann2011fastCost} & $97.52$ & $5.78$ & $93.94$ & $65.74$ \\ \hline
  Initialization & $9.33$ & $6.34$ & $5.62$ & $7.09$\\  
  Our upsampling & $9.96$ & $6.54$ & $\underline{5.52}$ & $7.34$\\  
Kopf \emph{et al.}~\cite{Kopf2007jointBilateral} & $10.23$ & $7.45$ & $5.68$ & $7.78$\\ \hline
    Dal Mutto \emph{et al.}~\cite{Mutto2012locallyConsistent} & $\bf{3.76}$ & $6.56$ & $8.69$ & $\underline{6.34}$ \\ 
EPC~\cite{Gandhi12} & $8.54$ & $6.61$ & $5.72$ & $6.95$\\ 
  F-MRF~\cite{Zhu08} & $8.96$ & $\underline{4.67}$ & $6.18$ & $6.67$\\ 
  F-ECC & $\underline{6.98}$ & $\bf{4.19}$ & $\bf{5.39}$ & $\bf{5.52}$\\ \hline
\end{tabular}
\label{table:mutto_dataset}
\end{table}
Dal Mutto~\emph{et al.}~\cite{Mutto2012locallyConsistent} provide real TOF-stereo data along with ground truth disparities, shown in Fig.~\ref{fig:mutto_scenes}. To be consistent with~\cite{Mutto2012locallyConsistent}, we upsample the depth in a similar way, using a bilateral filter, which benefits from color segmentation. This procedure is used to initialize the depth in \emph{all} fusion schemes. Table~\ref{table:mutto_dataset} shows the mean square error (MSE) of the disparity estimation for several algorithms. The contribution of the stereo data in fusion is unquestionable in scenes A and B, where the depth varies locally. All of the fusion methods obtain a more accurate map than the initial one. However, scene C contains only planar objects, and the upsampling methods provide good results. Although the proposed scheme does not perform best in all examples, it always improves the initial estimate, which demonstrates the advantage of adaptive fusion (very similar numbers are obtained with F-EMCC). As expected, the use of stereo data only (e.g.,~\cite{Rhemann2011fastCost}) performs well only with the textured scene B. Our upsamping filter is more accurate than~\cite{Kopf2007jointBilateral} and less accurate than the filter proposed by~\cite{Mutto2012locallyConsistent}. Note that our implementation of \cite{Kopf2007jointBilateral} achieves better results than those reported in~\cite{Mutto2012locallyConsistent}. 

\begin{figure}
\centering
\includegraphics[width=3.70cm, height=2.70cm]{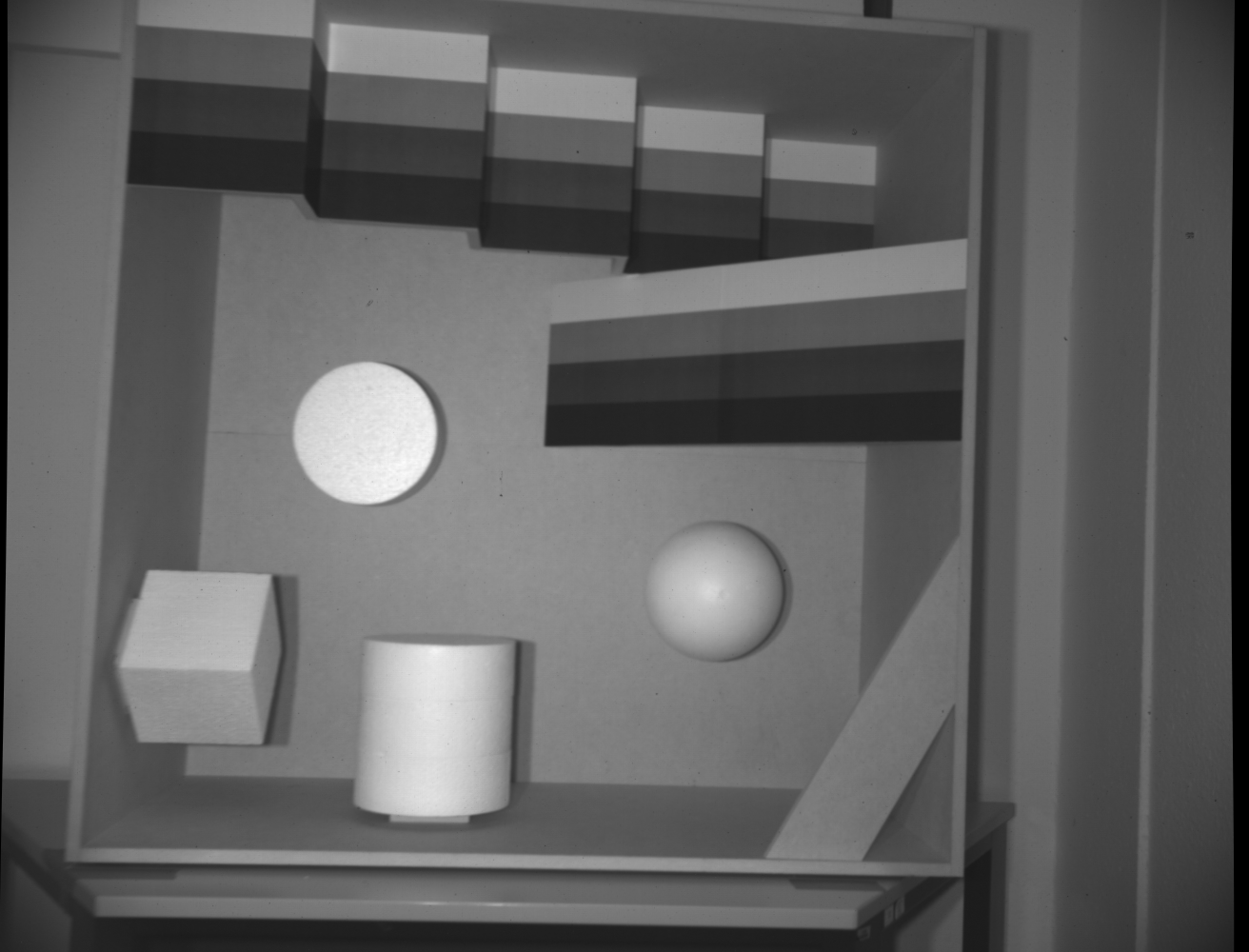}
\includegraphics[width=3.70cm, height=2.70cm]{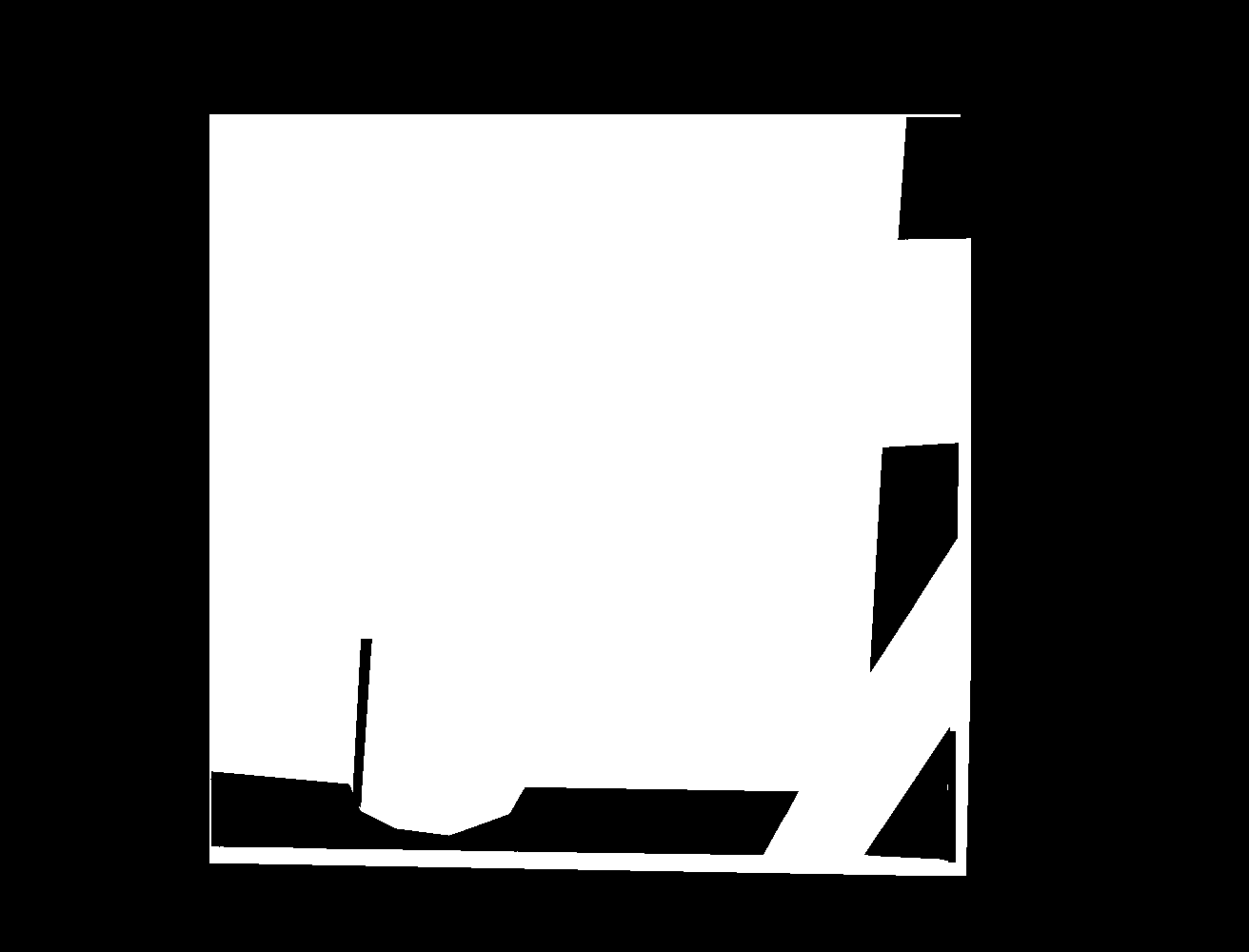}
\caption{The left image(left) and the mask that excludes inter-reflections (right) of the HCIbox dataset~\cite{Nair2012}.}
\label{fig:hcibox}
\vskip -0.4cm
\end{figure}

We also assess the performance of the fusion methods on the HCIbox data-set~\cite{Nair2012}. The scene shows the interior of a box that contains some objects (Fig.~\ref{fig:hcibox}). Note that there is no texture, apart from some horizontal lines on the stairs and the ramp, hence stereo methods tend to fail. We follow the experimental setup of~\cite{Nair2012}, thus evaluating the depth estimation based on some statistics of the absolute error, after excluding inter-reflection areas (see Fig.~\ref{fig:hcibox}). We do not include the results of~\cite{Nair2012}, since the authors provided a different inter-reflection mask with larger support area than the one used in~\cite{Nair2012}.\footnote{Personal communication with R. Nair.} Table~\ref{table:hci_dataset} shows the error statistics of the algorithms. All of the fusion methods start from the same initial map, obtained by our upsampling method. The proposed fusion method achieves the lowest mean and median error (similar results are obtained with F-EMCC).  The variance of F-MRF is increased (a local bias was observed owing to the global smoothness), while its median remains low. Because of the depth discontinuities, \cite{Kopf2007jointBilateral} yields a less accurate result compared to our upsampling.

\begin{table}[b!]
\centering
\caption{Evaluation on HCIBox dataset~\emph{et al.}~\cite{Nair2012}}
\begin{tabular}{|l|c|c|c|c|c|}

   \hline
  ~& Mean & St.\,D. & $1^{st}$\,Quart.  & Median & $3^{rd}$\,Quart. \\ \hline
  Kopf \emph{et al.}~\cite{Kopf2007jointBilateral} & $3.23$ & $\bf{4.00}$ & $0.93$ & $2.15$ & $3.85$ \\ 
  Initialization & $3.00$ & $\underline{4.21}$ & $0.84$ & $1.94$ & $3.61$ \\ \hline
  EPC~\cite{Gandhi12} & $3.01$ & $4.32$ & $0.90$ & $1.93$ & $\underline{3.17}$ \\ 
  F-MRF~\cite{Zhu08} & $\underline{2.95}$ & $4.96$ & $\underline{0.83}$ & $\underline{1.85}$ & $\underline{3.17}$ \\ 
  F-ECC & $\bf{2.53}$ & $4.25$ & $\bf{0.62}$ & $\bf{1.38}$ & $\bf{2.64}$ \\ \hline
\end{tabular}
\label{table:hci_dataset}
\end{table}

\begin{figure*}[htb]
\centering
\includegraphics[height=20.3mm, width=28mm]{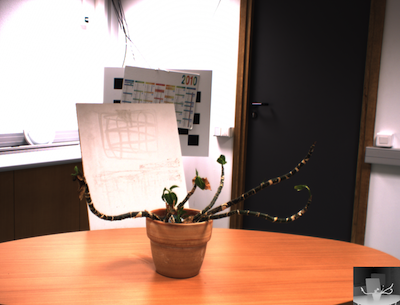}
\includegraphics[height=20.3mm, width=28mm]{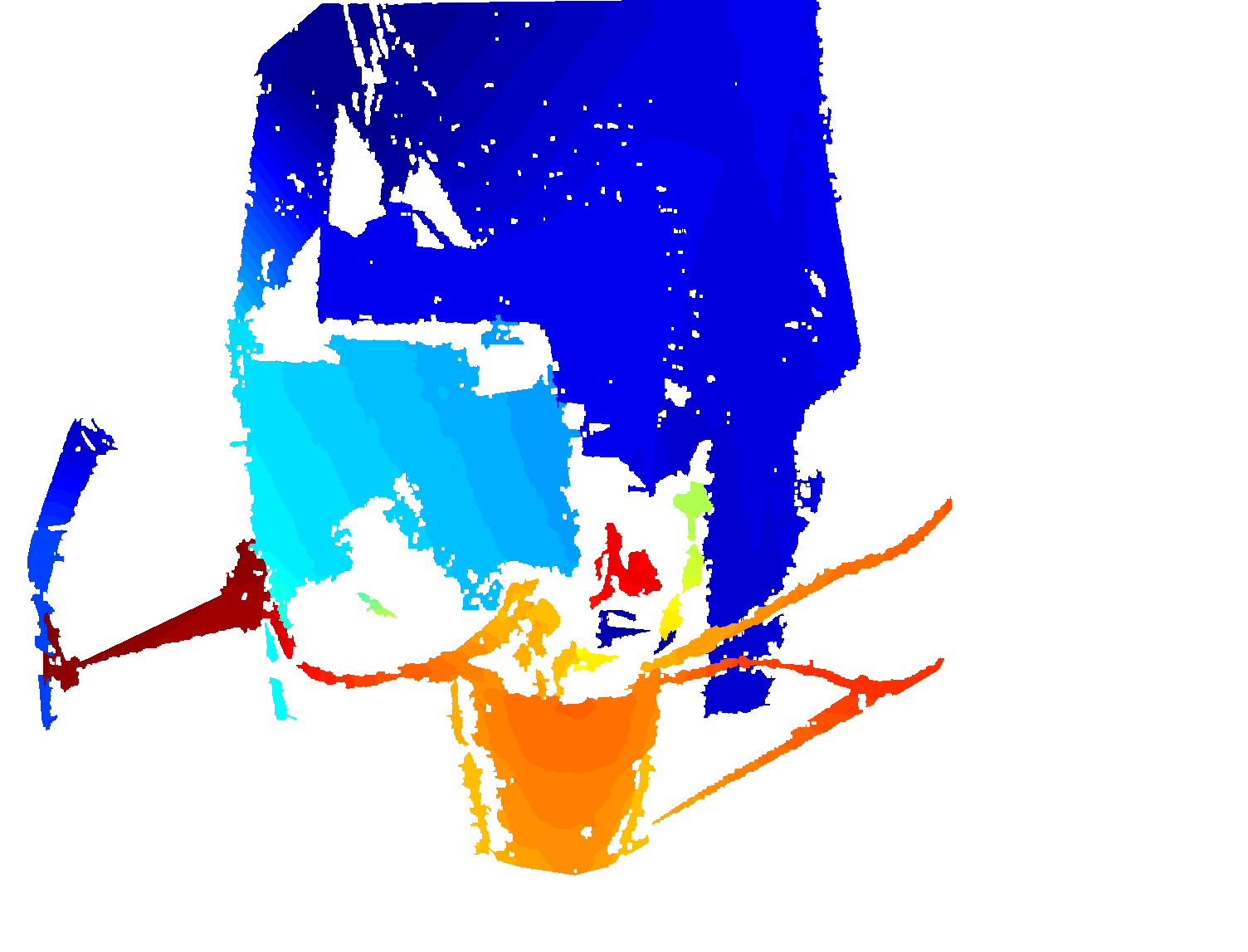}
\includegraphics[height=20.3mm, width=28mm]{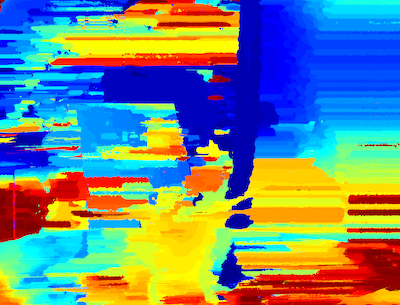}
\includegraphics[height=20.3mm, width=28mm]{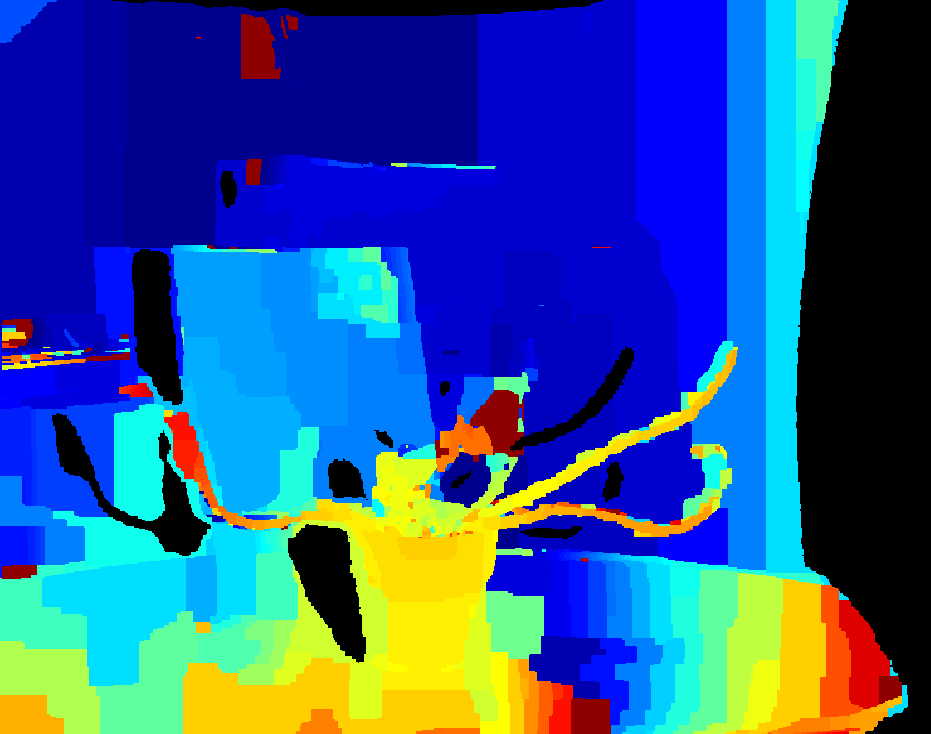}
\includegraphics[height=20.3mm, width=28mm]{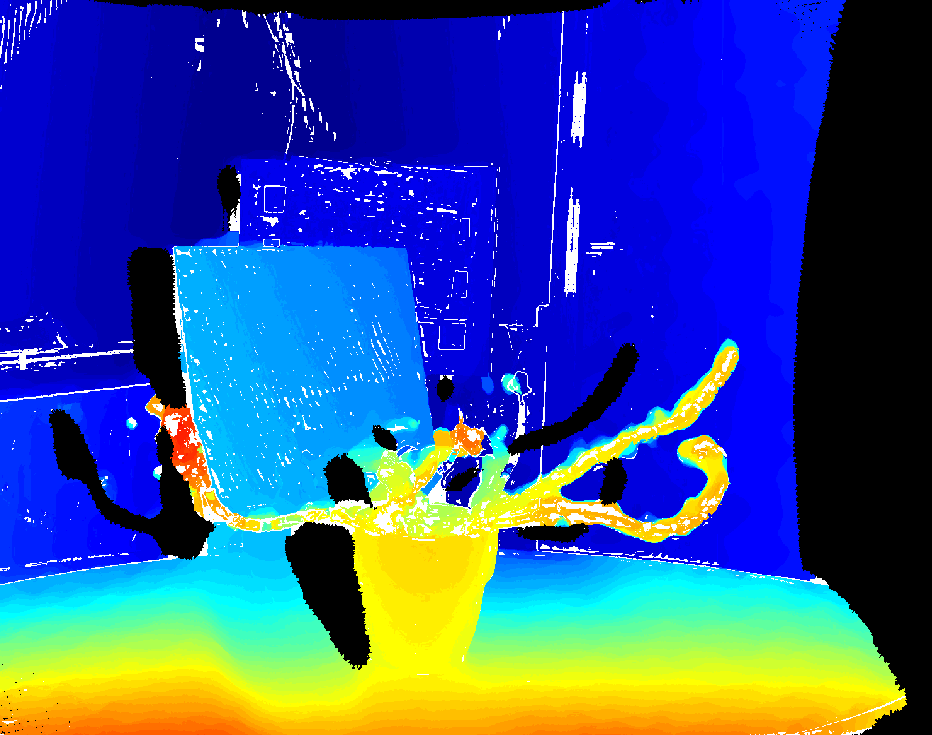}
\includegraphics[height=20.3mm, width=28mm]{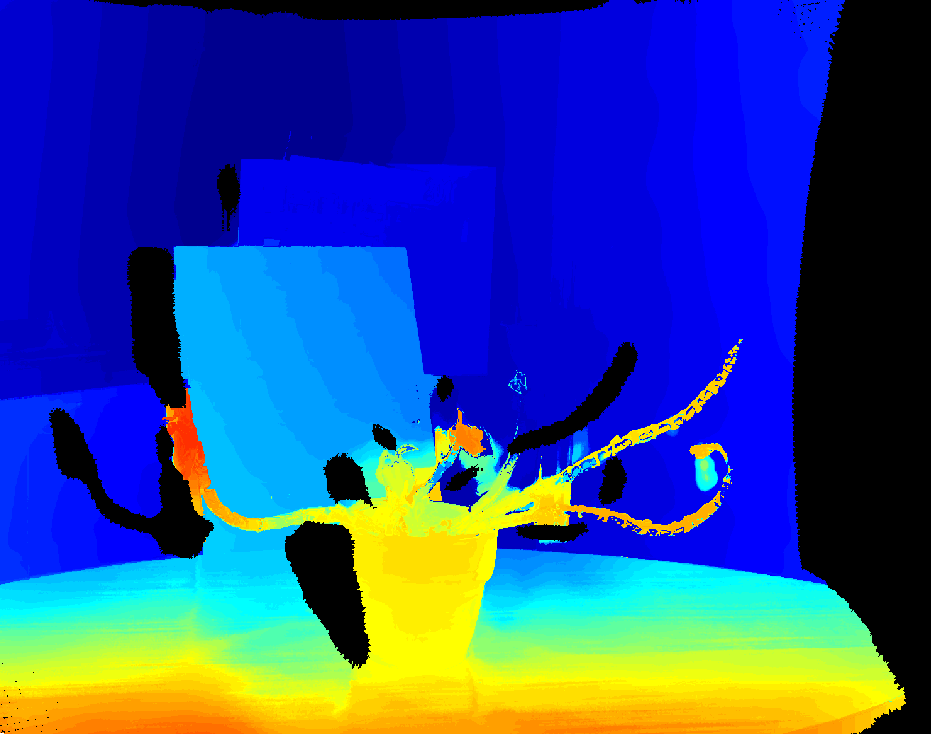}

\includegraphics[height=20.3mm, width=28mm]{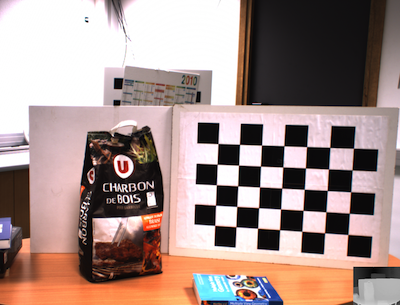}
\includegraphics[height=20.3mm, width=28mm]{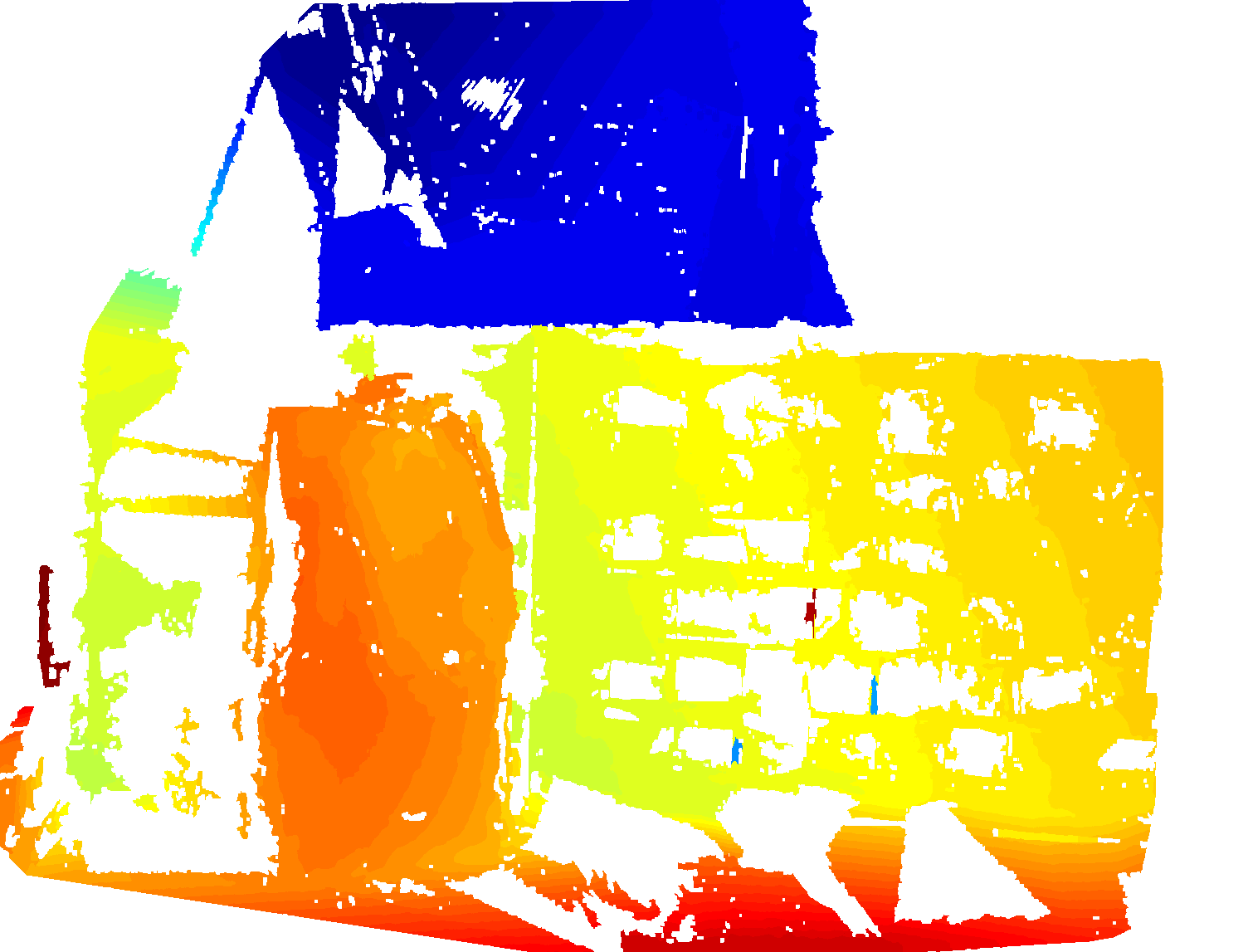}
\includegraphics[height=20.3mm, width=28mm]{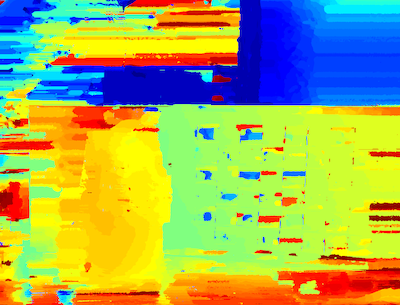}
\includegraphics[height=20.3mm, width=28mm]{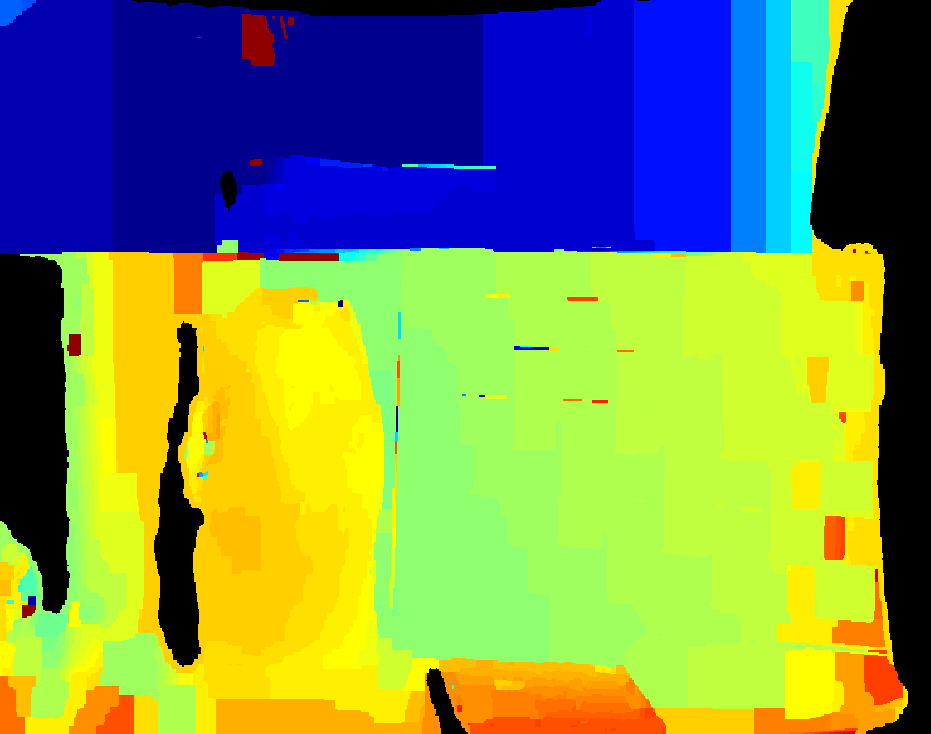}
\includegraphics[height=20.3mm, width=28mm]{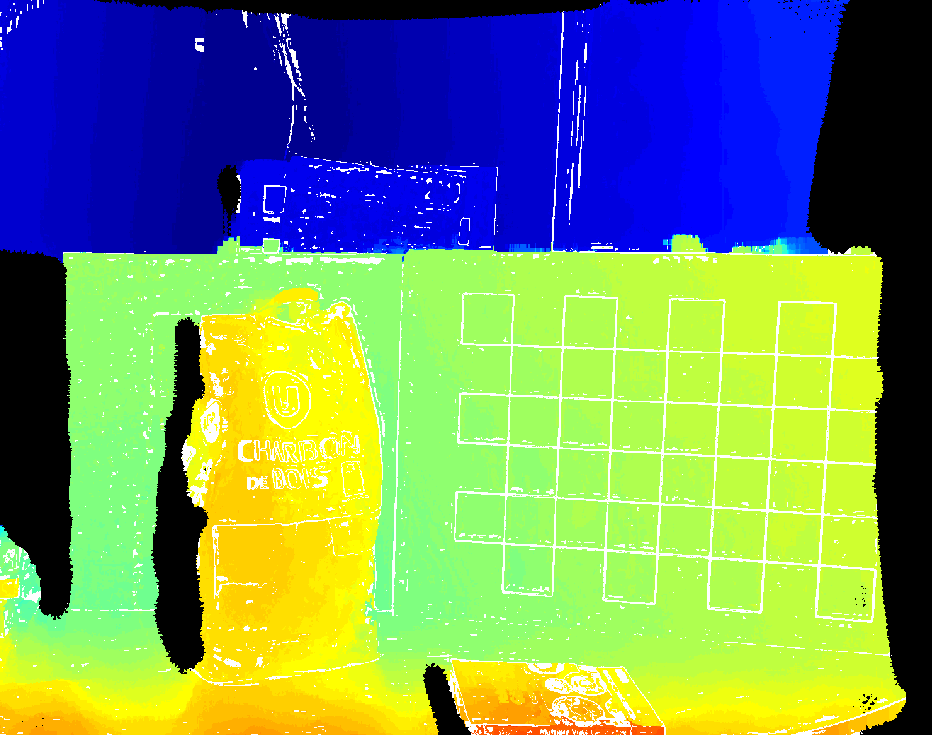}
\includegraphics[height=20.3mm, width=28mm]{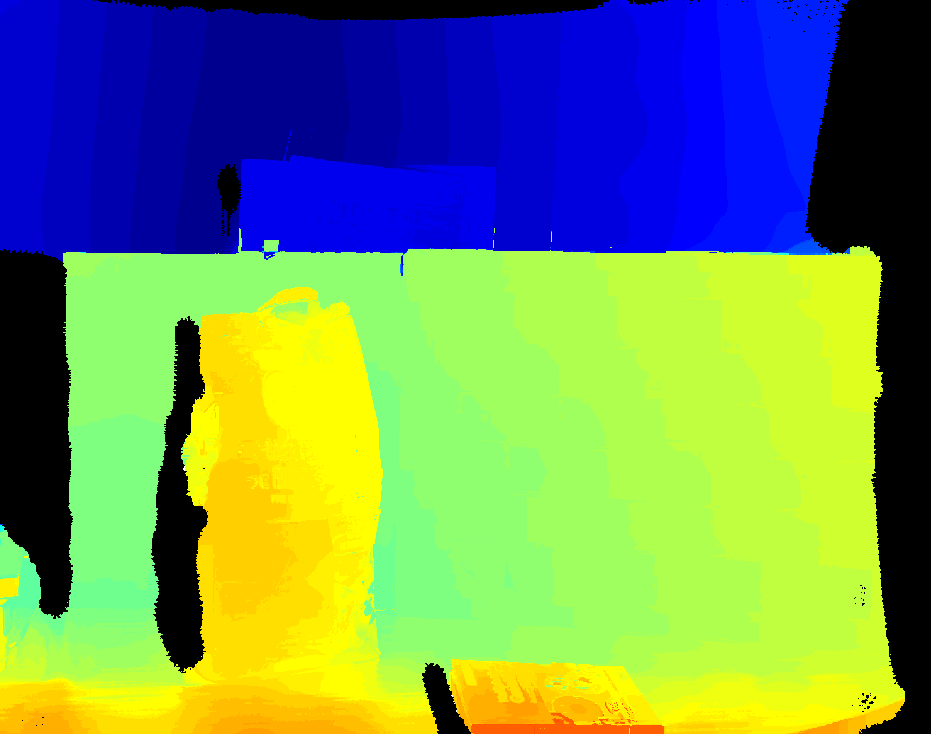}

\includegraphics[height=20.3mm, width=28mm]{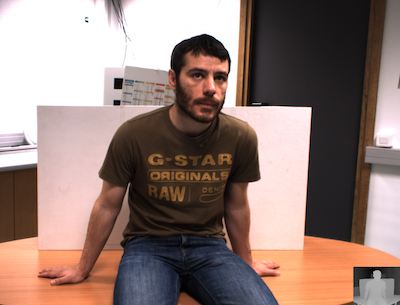}
\includegraphics[height=20.3mm, width=28mm]{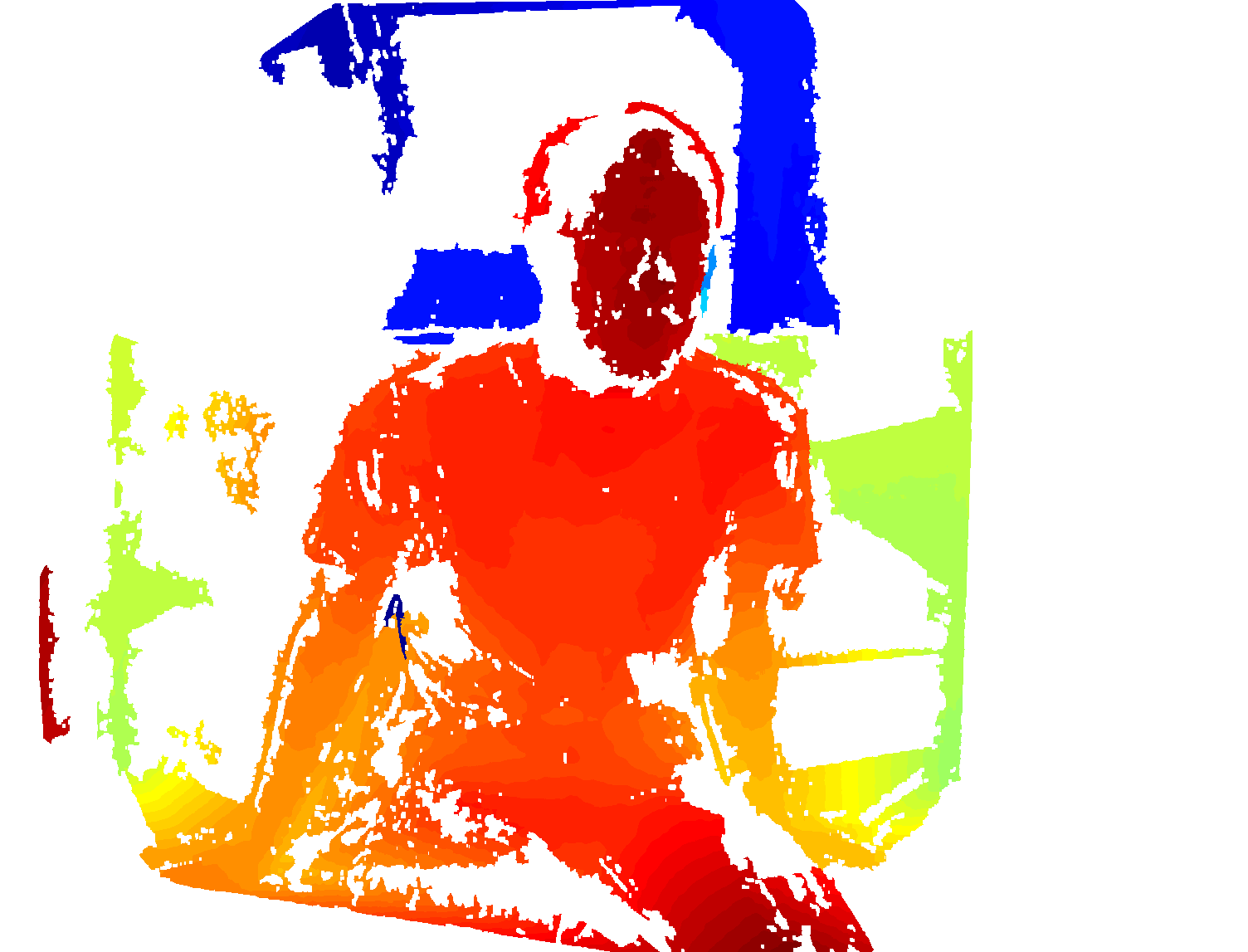}
\includegraphics[height=20.3mm, width=28mm]{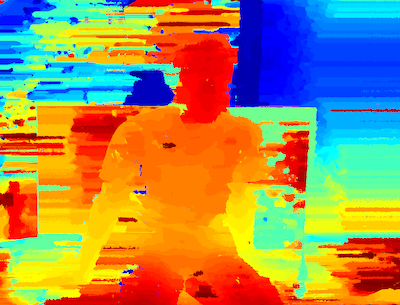}
\includegraphics[height=20.3mm, width=28mm]{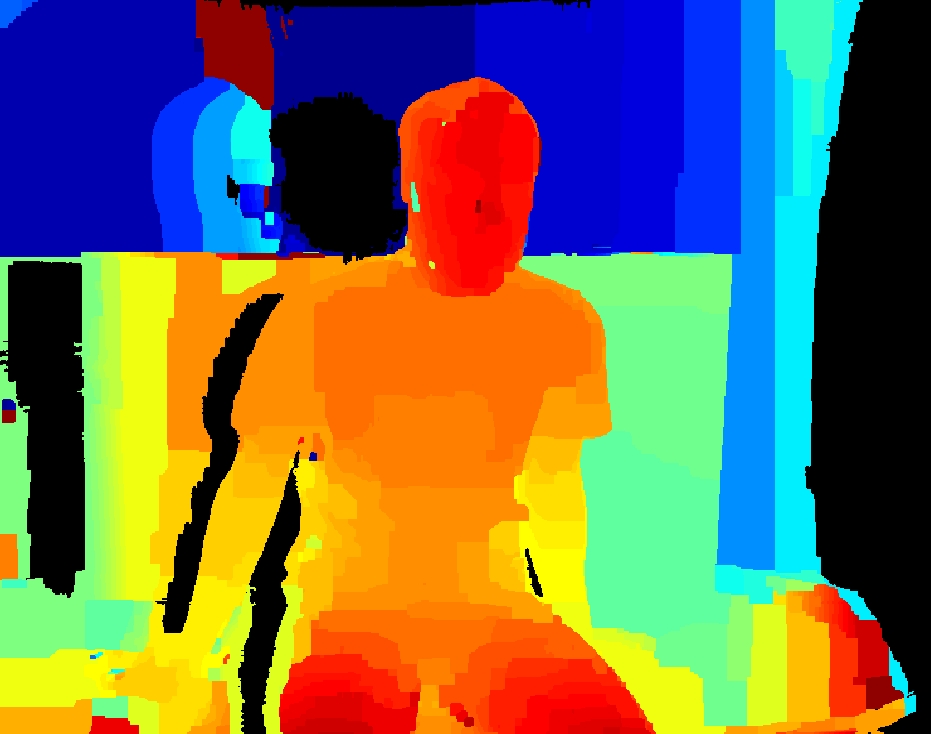}
\includegraphics[height=20.3mm, width=28mm]{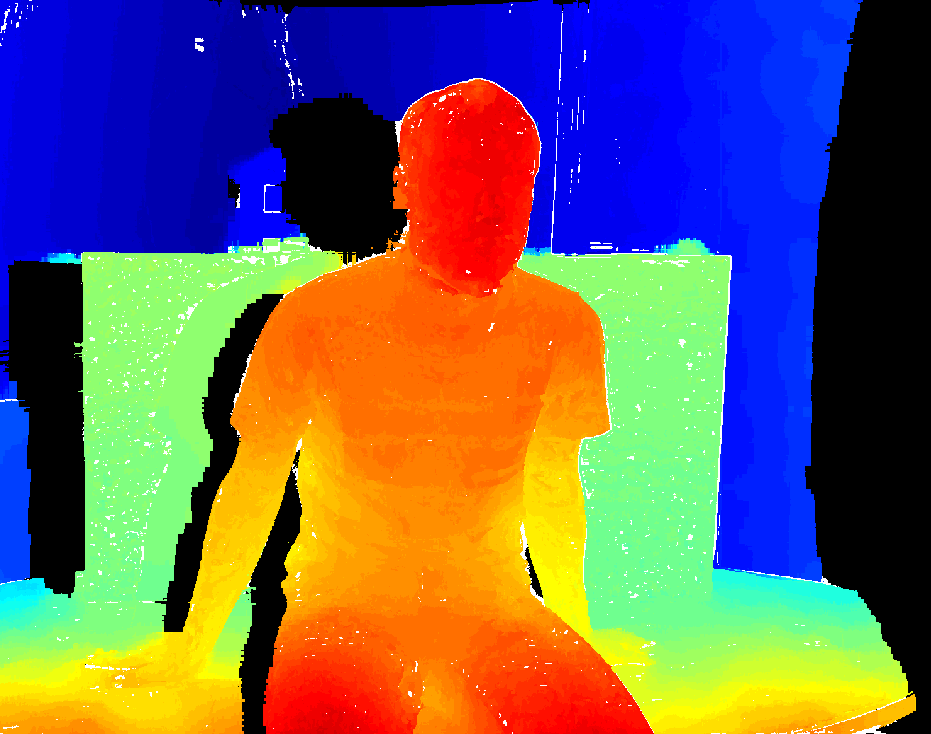}
\includegraphics[height=20.3mm, width=28mm]{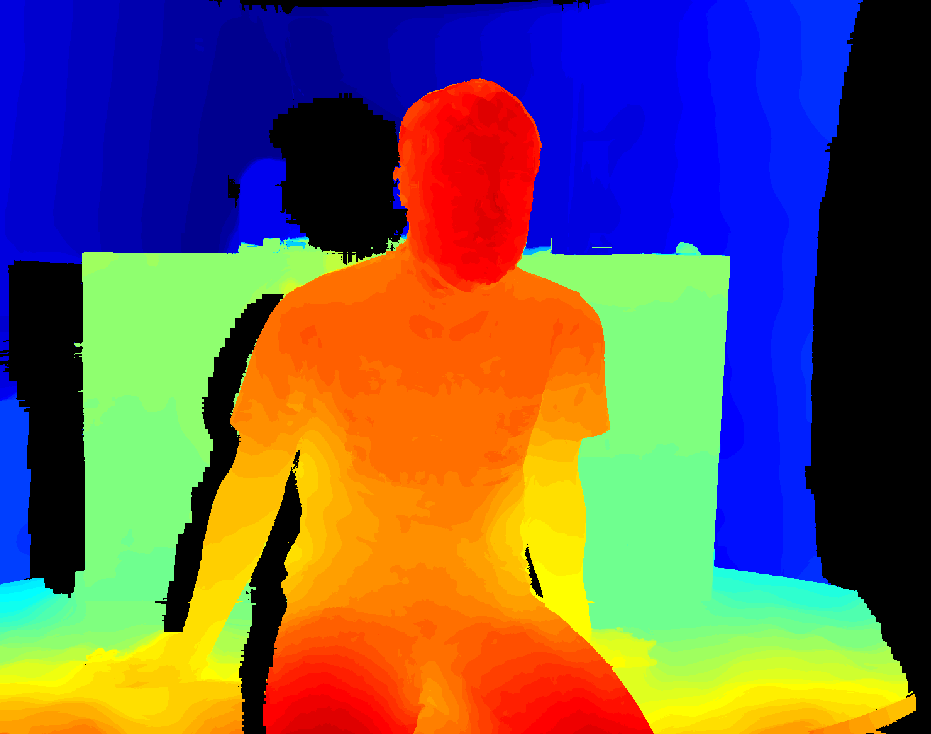}

\includegraphics[height=20.3mm, width=28mm]{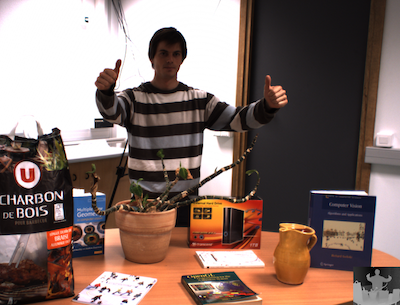}
\includegraphics[height=20.3mm, width=28mm]{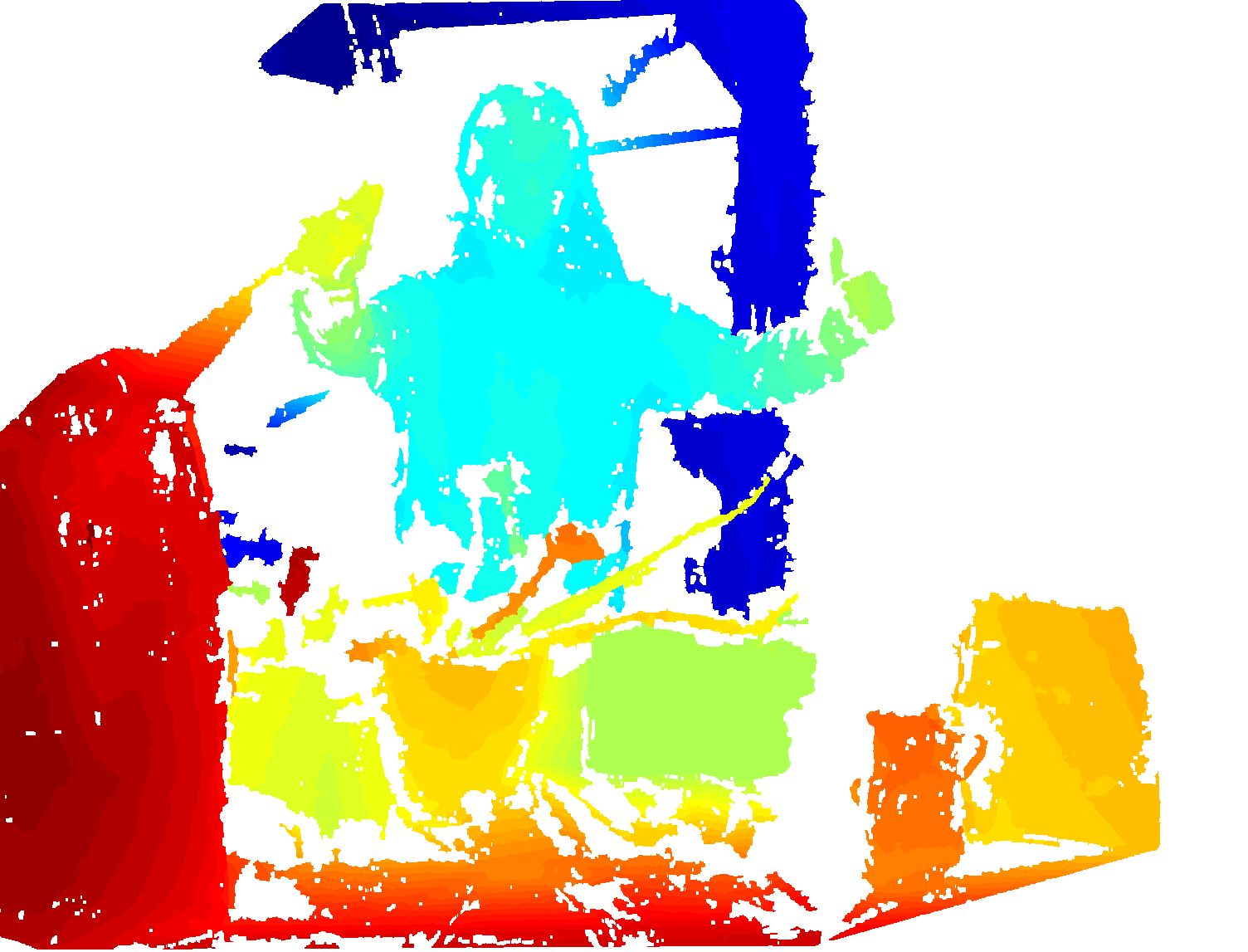}
\includegraphics[height=20.3mm, width=28mm]{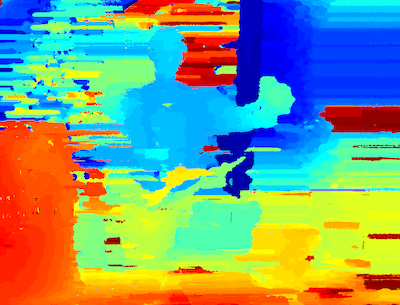}
\includegraphics[height=20.3mm, width=28mm]{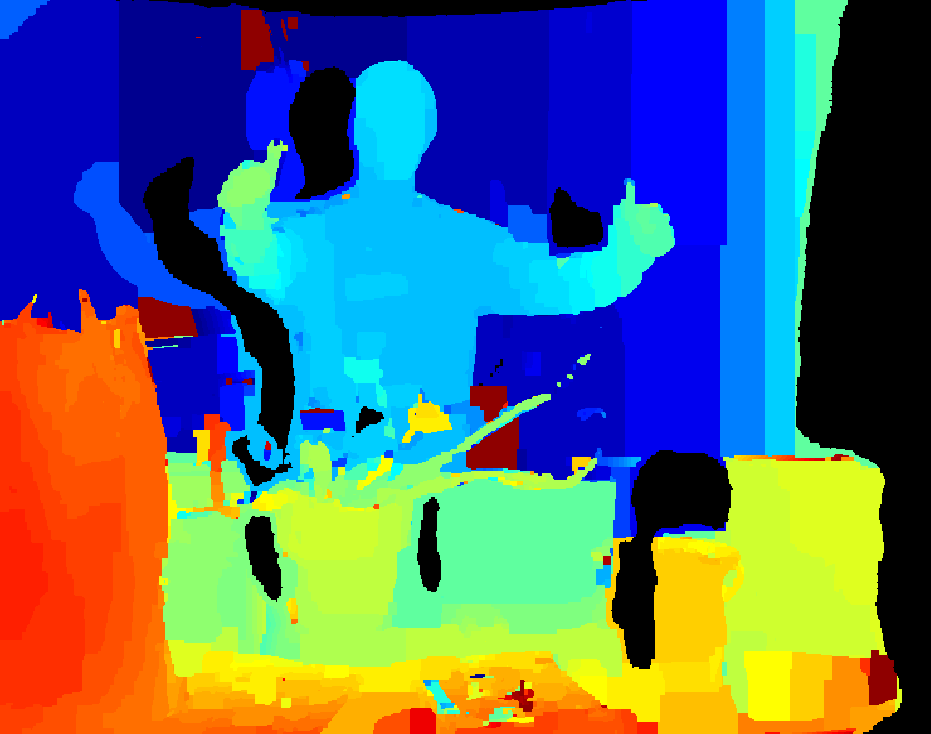}
\includegraphics[height=20.3mm, width=28mm]{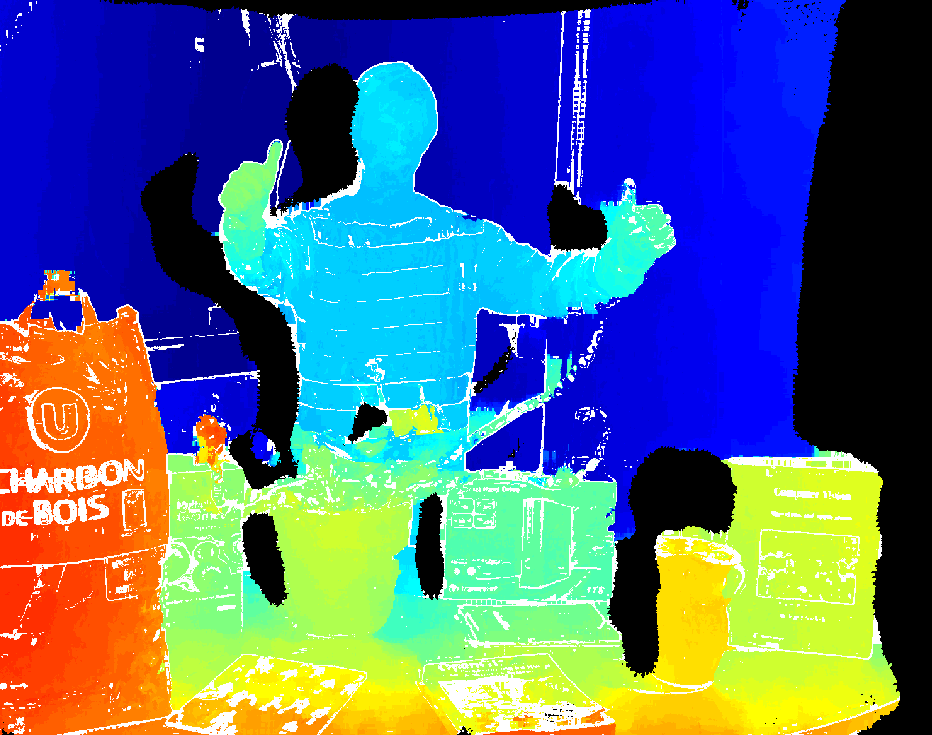}
\includegraphics[height=20.3mm, width=28mm]{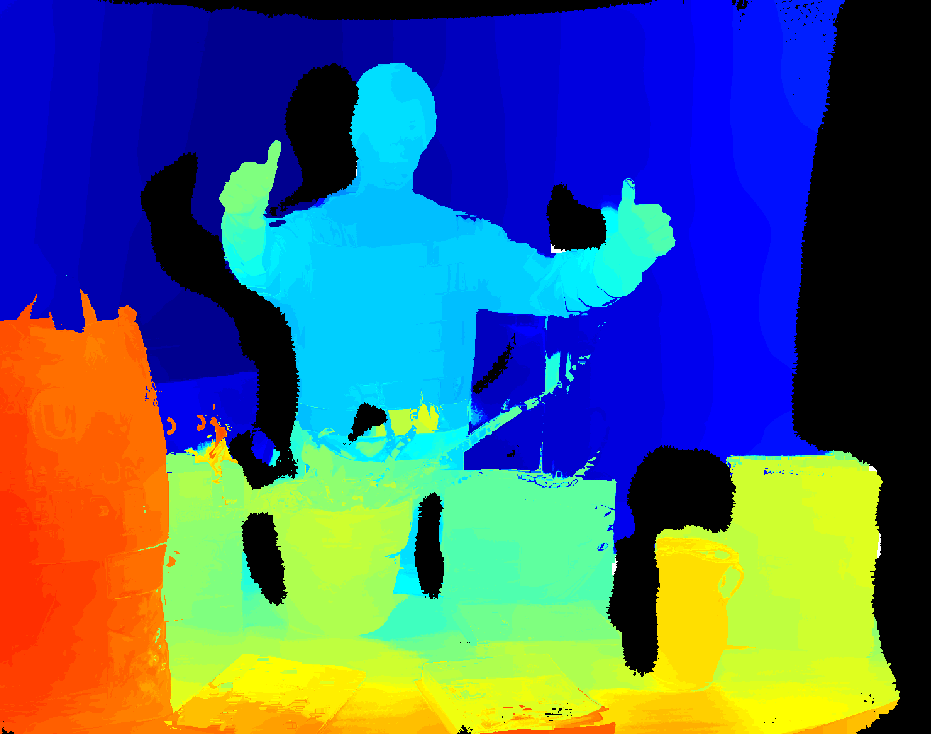}

\includegraphics[height=20.3mm, width=28mm]{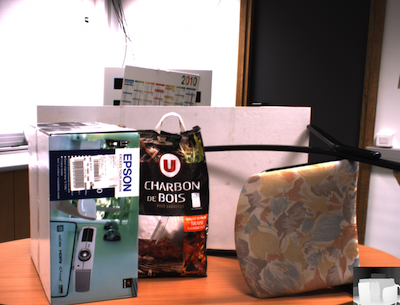}
\includegraphics[height=20.3mm, width=28mm]{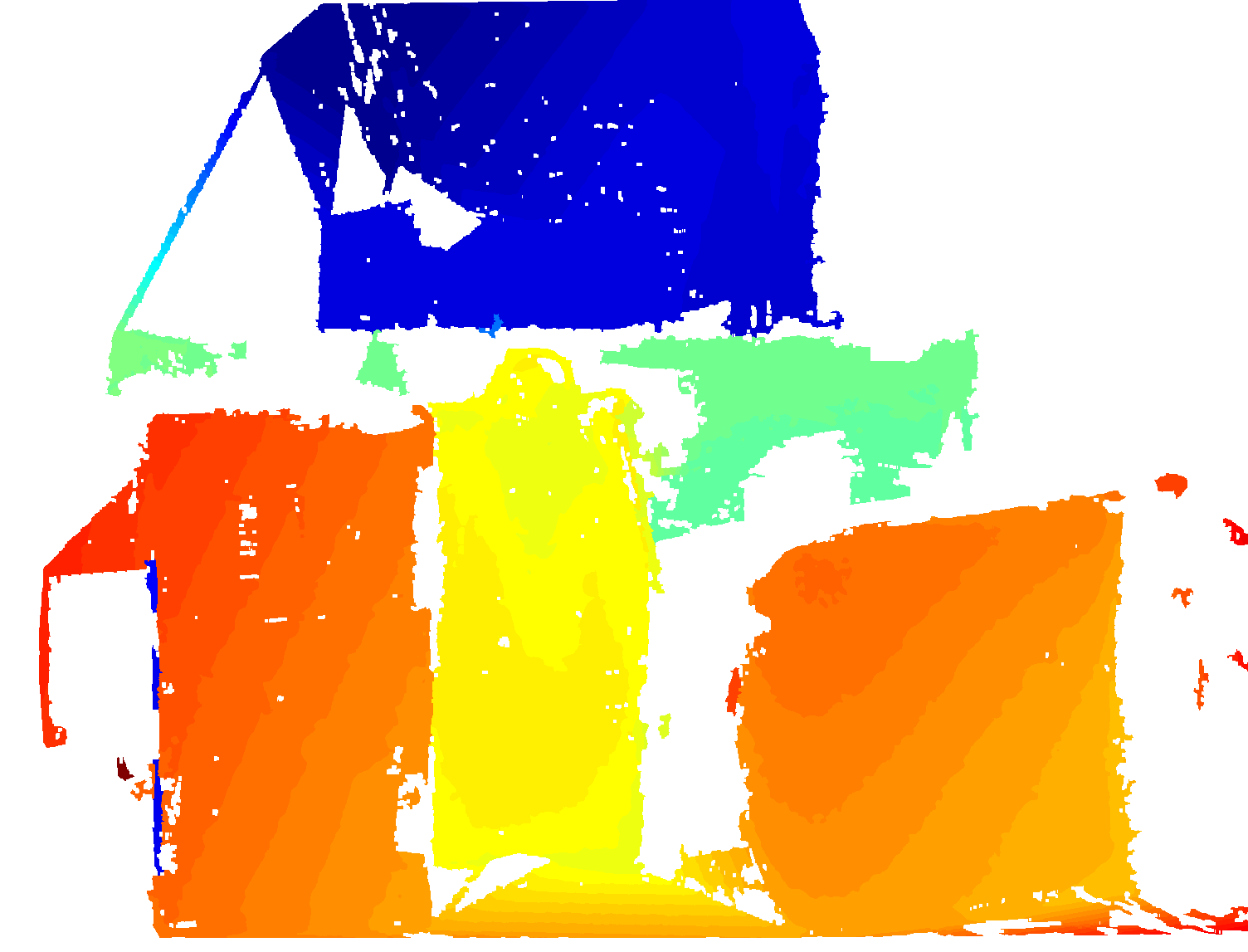}
\includegraphics[height=20.3mm, width=28mm]{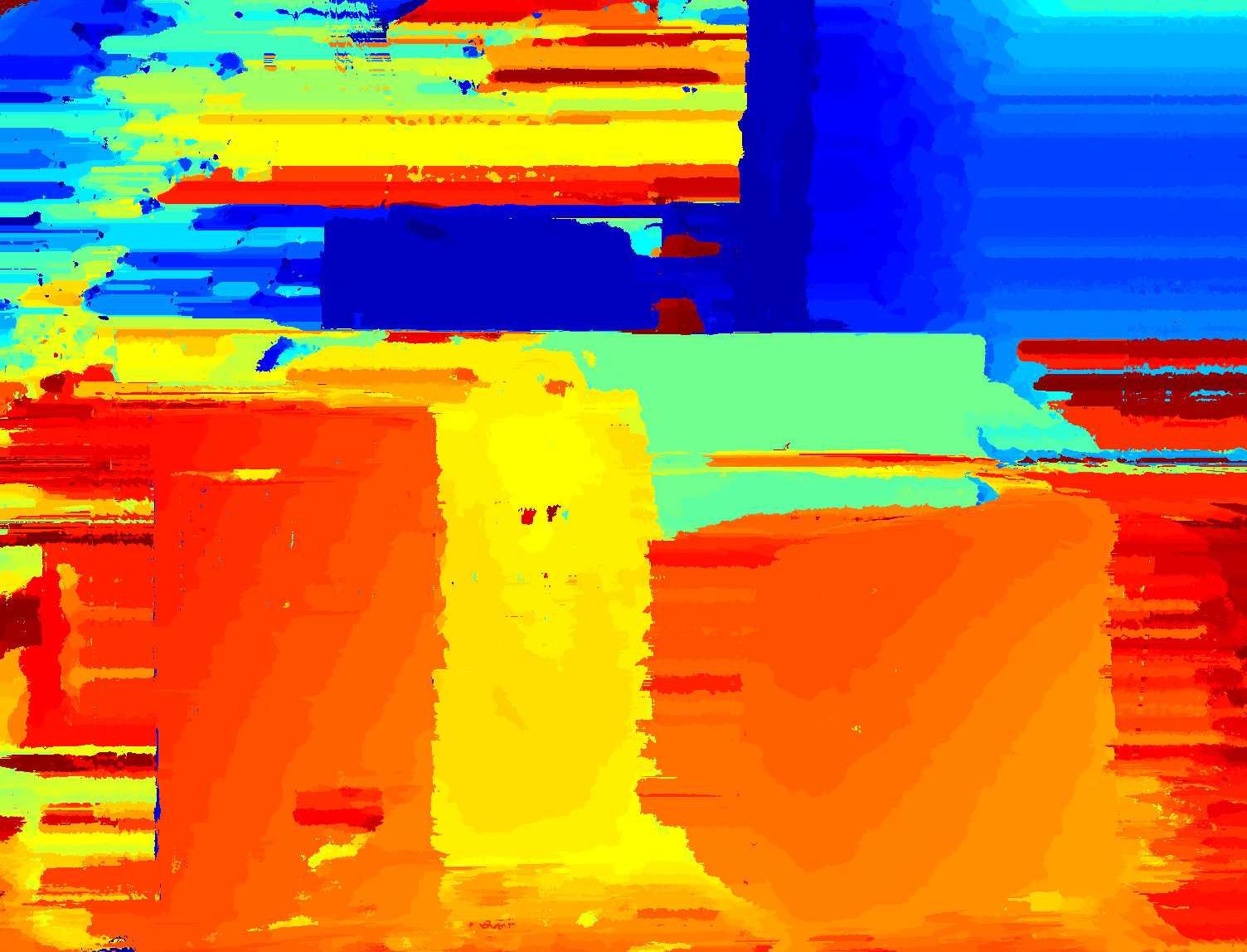}
\includegraphics[height=20.3mm, width=28mm]{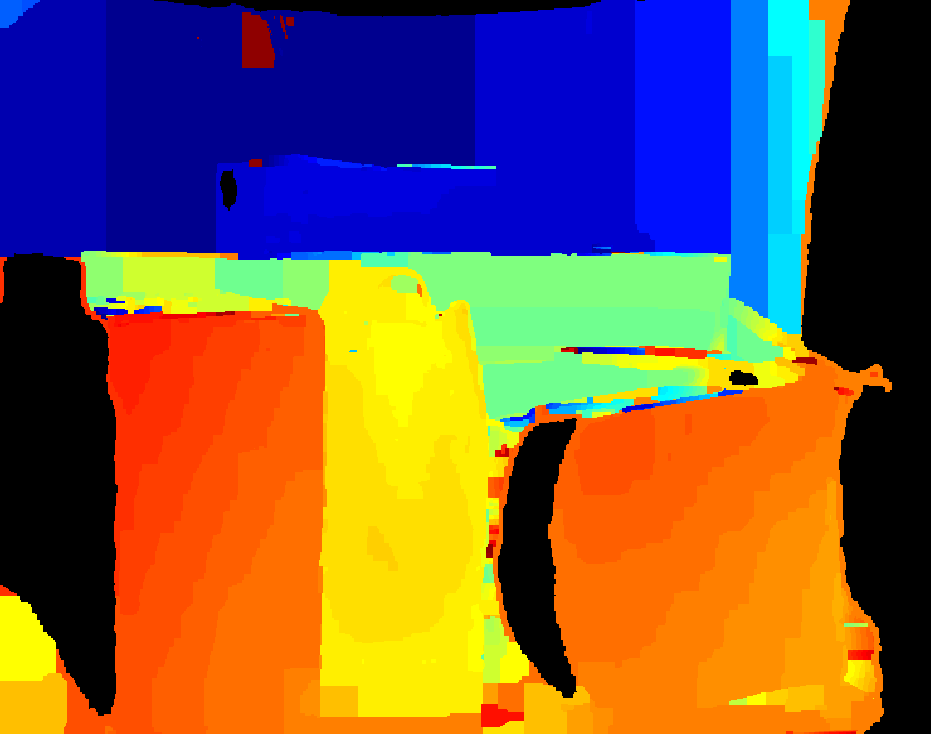}
\includegraphics[height=20.3mm, width=28mm]{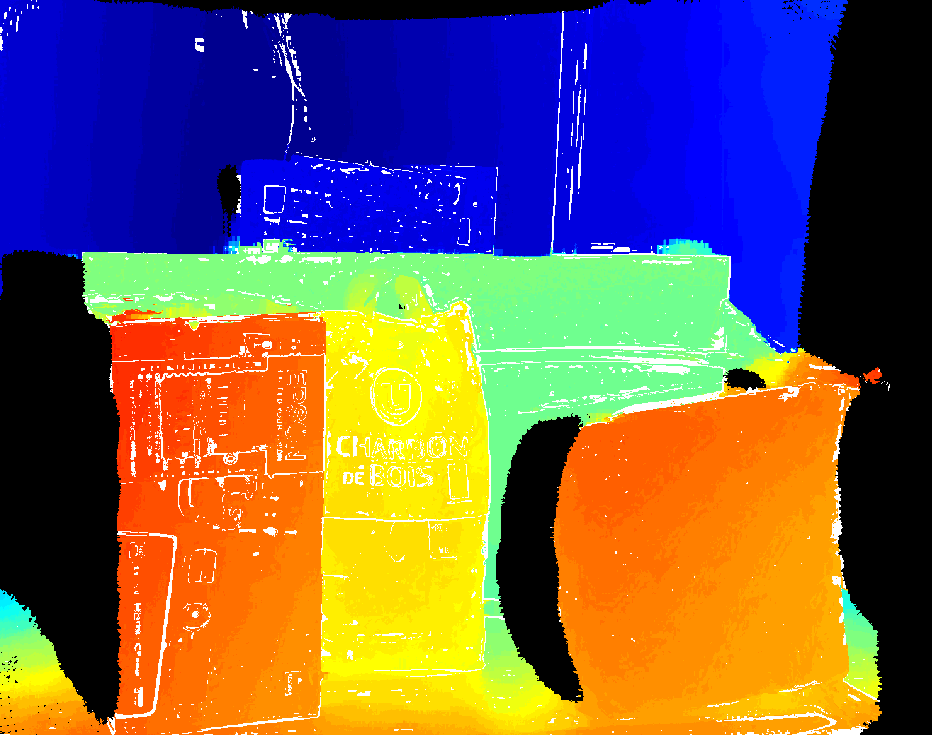}
\includegraphics[height=20.3mm, width=28mm]{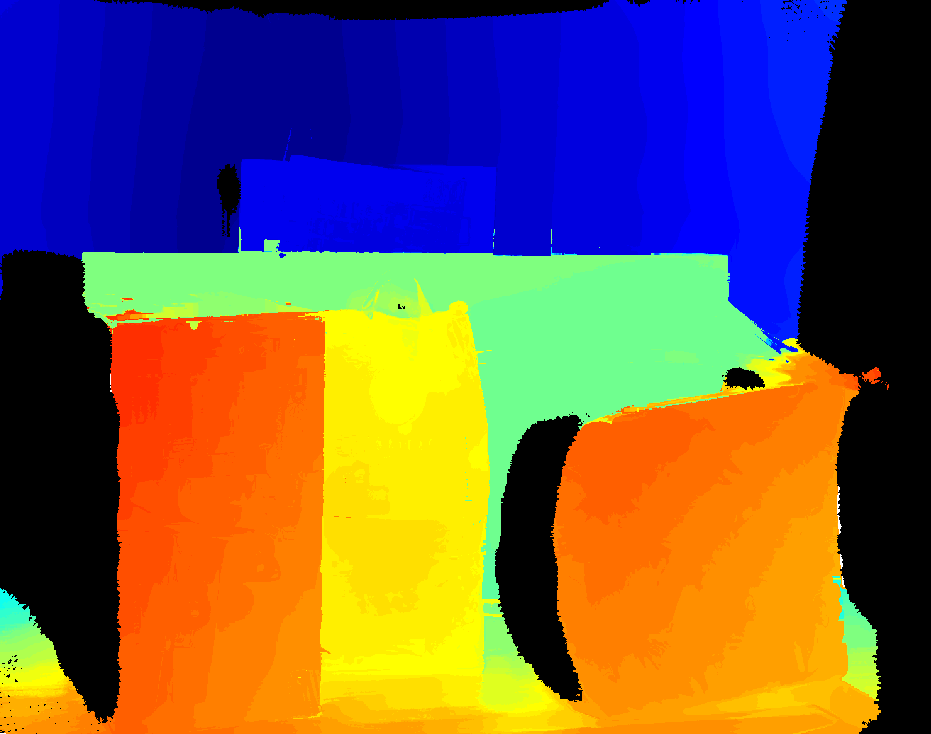}

\caption{HR images and disparity maps for MIXCAM dataset obtained by (from left to right) ELAS, FastAgg, F-MRF (GC), F-ECC, and F-ECC after post-processing.}
\label{fig:real_scenes}
\vskip -0.5cm
\end{figure*}

We also captured our own challenging TOF-stereo data-set using a synchronized camera setup, developed in collaboration with 4D View Solutions\footnote{\url{http://www.4dviews.com}}. Two HR ($1624\times 1224$) color cameras and a MESA Imaging SR4000 TOF (range) camera ($176\times 144$) are mounted on a rail. The stereo baseline is $50$cm approximately while the TOF camera is mounted in the middle. The algorithm of~\cite{Hansard2014} allows us to transform the depth-sensor measurements into a very sparse stereo disparity map ($1\%$ sparsity), with an average error of $1.0$ pixel. 
The sparse map is refined and upsampled as discussed in Sec.~\ref{sec:depth_initialization} (see also Fig.\ref{fig:upsampling}). For a fully real-time upsampling, we use our GPU-based implementation of the triangulation-based interpolation~\cite{Gandhi12}. Note that our stereo rectification puts the principal points in the same position, rather than making the optical axes parallel; this maximizes the overlap between the images, given the relatively wide baseline.

Our `MIXCAM' data-set contains challenging cases, e.g., periodicities, weakly textured areas, thin objects, depth discontinuities, and so on. The fusion algorithm merges the stereo and depth data and the outcome is a dense HR disparity map. We reuse our upsampling filter in a post-processing step to fill missing disparities. A streak-based method fills any remaining gaps in all algorithms. The results are shown in Fig.~\ref{fig:real_scenes}. White areas denote unmatched pixels, while black areas mark the detected TOF-occlusions. The left column shows the left image, with the TOF image shown in the bottom-right corner at the true scale. Next columns show the results of ELAS, FastAgg, F-MRF (GC) and F-ECC algorithms; the last column show the disparity maps of F-ECC after post-processing. 

ELAS fills local areas, surrounded by textured points, through an interpolation scheme. We intentionally show the results of ELAS before the streak-based filling; as opposed to FastAgg, where missing disparities after the left-right consistency check are filled. Clearly, a pure stereo algorithm cannot deal with large untextured areas, and the post-filling is unreliable. F-MRF provides fully dense results. Note that we run F-MRF with half-resolution images ($812\times 612$), due to its tremendous memory requirements. Moreover, we set a fixed value along the disparity range for the data-term of all TOF-occlusion points, so that the global inference becomes independent of this area. F-MRF provides artifacts in stereo occlusions, that are next to TOF-occlusions when the scene contains large foreground objects. As with~\cite{Zhu11}, the results of F-MRF scheme verify the lack of an adaptive fusion of the depth- and stereo-consistency data terms, as opposed to our methods. However, F-MRF seems to deal better with very thin objects (e.g. the branch of the plant), as already discussed above. Note that the biased range measurements of very slanted surfaces (e.g. the table-top) negatively affect the fusion schemes, in particular when the table surface lacks texture (e.g. first example). The proposed scheme provides very good results on average, especially after the post-processing step, which fills the gaps and refines the disparities. We obtain very similar results with the F-EMCC method, while EPC provides results \emph{visually} close to ours, but with more gaps. The bilateral upsampling of~\cite{Kopf2007jointBilateral} provides visually good results, but with blurred depth discontinuities (see also Fig.~\ref{fig:upsampling}).

\section{Conclusions}\label{sec:conclusions}
We have presented a high-resolution stereo matching algorithm that is guided by low-resolution depth data, thus helping the algorithm to compensate for its difficulty in estimating disparities over weakly textured areas. We cast the problem into a MAP formulation whose inference is obtained through a series of local optimization problems, solved hierarchically in a seed-growing manner. The latter characteristic yields an intrinsically efficient solution that allows for near real-time matching of 2.0MP images. The data-term of the energy function benefits from a correlation function that is capable of providing scores at subpixel disparities, from an adaptive cost aggregation step inside the window based on the depth data, and from an adaptive fusion of stereo- and depth-consistency terms based on the scene texture and the camera geometry. These properties lead to a more selective growing process that prevents
the algorithm from propagating incorrect disparities. As a result, a low-complexity method builds an accurate high-resolution disparity map. A quantitative comparison against pure stereo and stereo-depth fusion algorithms, as well as a qualitative assessment on real data, has validated the strong performance of the proposed method. Future research will include the optimum visiting order for seeds in the growing framework, as well as an adaptive window size, based on the local surface orientation.

\bibliographystyle{IEEEtran}

\vskip -0.5cm

\begin{IEEEbiography}[{\includegraphics[width=1in,height=1.25in,clip,keepaspectratio]{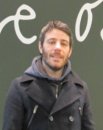}}]{Georgios D. Evangelidis} received his BSc, MSc and PhD degree in computer science in 2001, 2003 and 2008 respectively from the University of Patras, Greece. From 2007 to 2009 he was an adjunct lecturer of the Technological Institute of Larissa, Greece. During 2009-2010, he was an ERCIM (Alain Bensoussan) Fellow and joined the Fraunhofer Institute for Intelligent Analysis and Information Systems (IAIS) in Sankt Augustin, Germany, as a postdoctoral researcher. Currently, he is a researcher at the Perception Team of INRIA Grenoble, France. His research interests are in the area of computer vision and include 3D reconstruction, Gesture recognition and Depth-Stereo Fusion.
\end{IEEEbiography}
\vskip -0.5cm

\begin{IEEEbiography}[{\includegraphics[width=1in,height=1.25in,clip,keepaspectratio]{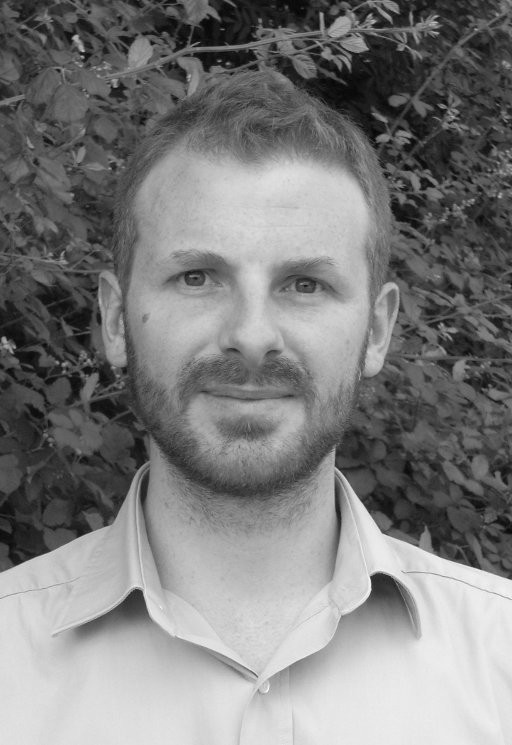}}]
{Miles Hansard} is a lecturer in Computer Science at Queen Mary, University of London. He is a member of the Vision Group, and of the QMUL Centre for Intelligent Sensing. His research interests include 3D scene modelling, depth cameras, and human vision. He has BSc, MRes and PhD degrees from University College London.
\end{IEEEbiography}
\vskip -0.5cm

\begin{IEEEbiography}[{\includegraphics[width=1in,
height=1.25in,clip,keepaspectratio]{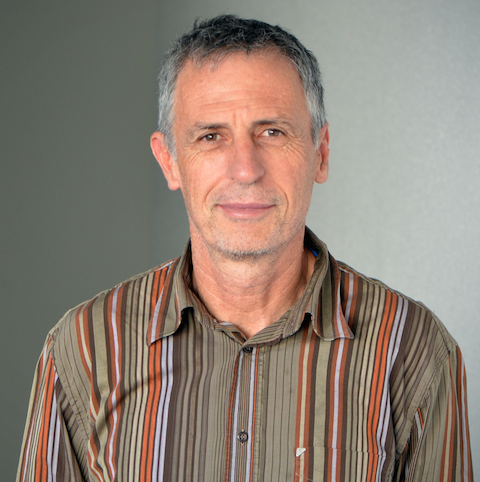}}]{Radu Horaud} 
received the B.Sc. degree in electrical engineering, the M.Sc. degree
in control engineering, and the Ph.D. degree in computer science from
the Institut National Polytechnique de Grenoble, Grenoble, France. 
Currently he holds a position of director of research with the Institut National de Recherche en Informatique et
Automatique (INRIA), Grenoble Rh\^one-Alpes, Montbonnot, France, where
he is the founder and head of the PERCEPTION team. His
research interests include computer vision, machine learning, audio signal processing, 
audiovisual analysis, and robotics. He is an area editor of the
\textit{Elsevier Computer Vision and Image Understanding}, a member of
the advisory board of the \textit{Sage International Journal of Robotics
  Research}, and an associate editor of the
\textit{Kluwer International Journal of Computer Vision}. He was
Program Cochair of the Eighth IEEE International Conference on
Computer Vision (ICCV 2001). In 2013, Radu Horaud was awarded a five year ERC Advanced Grant for his project \textit{Vision and Hearing in Action} (VHIA).
\end{IEEEbiography}


\end{document}